\documentclass[pdflatex,sn-mathphys-num,iicol]{sn-jnl}

\usepackage{graphicx}%
\usepackage{multirow}%
\usepackage{amsmath,amssymb,amsfonts}%
\usepackage{amsthm}%
\usepackage{mathrsfs}%
\usepackage[title]{appendix}%
\usepackage[table,usenames,dvipsnames]{xcolor}%
\usepackage{textcomp}%
\usepackage{manyfoot}%
\usepackage{booktabs,tabularx, colortbl}%
\usepackage{algorithm}%
\usepackage{algorithmicx}%
\usepackage{algpseudocode}%
\usepackage{listings}%

\usepackage{hyperref}
\hypersetup{
            breaklinks,
            colorlinks,
            linkcolor=blue,
            citecolor=blue,
            filecolor=blue,
            urlcolor=blue}
\usepackage[capitalise]{cleveref}
\usepackage[many]{tcolorbox}
\newtcolorbox{myboxi}[2][]{
  breakable,
  title=#1,
  colback=white,
  colbacktitle=white,
  coltitle=black,
  fonttitle=\itshape,
  bottomrule=-0.1pt,
  toprule=-0.2pt,
  leftrule=5pt,
  rightrule=0pt,
  titlerule=0pt,
  arc=0pt,
  boxsep=1mm,
  outer arc=0pt,
  colframe=#2,
}
\usepackage{enumitem}
\usepackage{caption}
\usepackage{array}
\usepackage{tikz}
\usepackage{pgfplots}
\pgfplotsset{compat=newest}
\usepgfplotslibrary{colormaps}
\usepackage{collcell}
\usepackage{listofitems}
\usepackage{afterpage}
\usepackage{adjustbox}
\usepackage{placeins}
\usepackage{xcolor,pifont}
\usepackage{xstring}

\newcommand{\etal}{\textit{et al}. }
\newcommand{\ie}{\textit{i}.\textit{e}.}
\newcommand{\eg}{\textit{e}.\textit{g}.}

\newcommand*\colourcheck[1]{%
  \expandafter\newcommand\csname #1check\endcsname{\textcolor{#1}{\ding{51}}}%
}
\newcommand*\colourcross[1]{%
  \expandafter\newcommand\csname #1cross\endcsname{\textcolor{#1}{\ding{55}}}%
}
\colourcheck{blue}
\colourcheck{green}
\colourcross{red}

\definecolor{darkgreen}{rgb}{0.0, 0.75, 0.0}

\definecolor{bananayellow}{rgb}{1.0, 0.88, 0.21}

\usepackage{multirow}

\definecolor{lowcolor}{HTML}{ef8a62}
\definecolor{centralcolor}{HTML}{f7f7f7}
\definecolor{highcolor}{HTML}{67a9cf}
\definecolor{same_attr}{HTML}{009900}
\definecolor{diff_attr}{HTML}{FE0000}

\pgfplotsset{%
  colormap={PiYG}{%
    rgb=(0.93725490196, 0.54117647058, 0.38431372549)%
    rgb=(0.96862745098, 0.96862745098, 0.96862745098)%
    rgb=(0.40392156862, 0.66274509803, 0.81176470588)%
  }%
}

\def\pgfplotsshowcolormap#1{%
    \pgfplotscolormapifdefined{#1}{\relax}{%
        \pgfplotsset{colormap/#1}%
    }%
    \pgfplotscolormaptoshadingspec{#1}{1cm}\result
    \def\tempb{\pgfdeclarehorizontalshading{tempshading}{0.2cm}}%
    \expandafter\tempb\expandafter{\result}%
    \pgfuseshading{tempshading}%
}

\newlength{\savedtabcolsep}
\newcommand{\midsepremove}{\aboverulesep=0mm \belowrulesep=0mm}%
\usepackage{listofitems}
\setsepchar{ }

\newcommand{\ApplyGradient}[1]{%
  \IfStrEq{#1}{-}{%
    \pgfmathsetmacro{\Val}{\MinVal}
  }{%
    \pgfmathsetmacro{\Val}{#1}
  }%
  \pgfmathsetmacro{\MiddleVal}{\MinVal + (\MaxVal - \MinVal) / 2}%
  \ifdim \Val pt > \MiddleVal pt%
      \pgfmathsetmacro{\PercentColor}{max(min(100.0*(\Val - \MiddleVal)/(\MaxVal - \MiddleVal), 100.0), 0.00)}%
      \edef\HeatCell{\noexpand\cellcolor{highcolor!\PercentColor!centralcolor}}%
      \HeatCell$#1$%
  \else
      \pgfmathsetmacro{\PercentColor}{max(min(100.0*(\MiddleVal - \Val)/(\MiddleVal - \MinVal), 100.0), 0.00)}%
      \edef\HeatCell{\noexpand\cellcolor{lowcolor!\PercentColor!centralcolor}}%
      \HeatCell$#1$%
  \fi%
}

\newcommand{\ApplyGradientReverse}[1]{%
  \IfStrEq{#1}{-}{%
    \pgfmathsetmacro{\Val}{\MaxVal}
  }{%
    \pgfmathsetmacro{\Val}{#1}%
  }%
  \pgfmathsetmacro{\MiddleVal}{\MinVal + (\MaxVal - \MinVal) / 2}%
  \ifdim \Val pt > \MiddleVal pt%
      \pgfmathsetmacro{\PercentColor}{max(min(100.0*(\Val - \MiddleVal)/(\MaxVal - \MiddleVal), 100.0), 0.00)}%
      \edef\HeatCell{\noexpand\cellcolor{lowcolor!\PercentColor!centralcolor}}%
      \HeatCell$#1$%
  \else
      \pgfmathsetmacro{\PercentColor}{max(min(100.0*(\MiddleVal - \Val)/(\MiddleVal - \MinVal), 100.0), 0.00)}%
      \edef\HeatCell{\noexpand\cellcolor{highcolor!\PercentColor!centralcolor}}%
      \HeatCell$#1$%
  \fi%
}

\newcolumntype{\C}[2]{>{\def\MinVal{#1}\def\MaxVal{#2}\collectcell\ApplyGradient}c<{\endcollectcell}}

\newcolumntype{\CR}[2]{>{\def\MinVal{#1}\def\MaxVal{#2}\collectcell\ApplyGradientReverse}c<{\endcollectcell}}

\theoremstyle{thmstyleone}%
\theoremstyle{thmstyletwo}%

\theoremstyle{thmstylethree}%

\begin{document}

\title[SEVERE++]{SEVERE++: Evaluating Benchmark Sensitivity in Generalization of Video Representation Learning}


\author*[1]{\fnm{Fida Mohammad } \sur{Thoker}}\email{fida.thoker@kaust.edu.sa}

\author[1]{\fnm{Letian} \sur{Jiang}}\email{letian.jiang@kaust.edu.sa}
\author[1]{\fnm{Chen} \sur{Zhao}}\email{chen.zhao@kaust.edu.sa}
\author[2]{\fnm{Piyush} \sur{Bagad}}\email{piyush@robots.ox.ac.uk}
\author[3]{\fnm{Hazel} \sur{Doughty}}\email{h.r.doughty@liacs.leidenuniv.nl}
\author[1]{\fnm{Bernard} \sur{Ghanem}}\email{Bernard.Ghanem@kaust.edu.sa}
\author[4]{\fnm{Cees G. M.} \sur{Snoek}}\email{c.g.m.snoek@uva.nl }


\affil[1]{\orgdiv{CEMSE}, \orgname{King Abdullah University of Science and Technology}, \orgaddress{\city{Makkah}, 
\country{KSA}}}

\affil[2]{\orgdiv{VGG}, \orgname{University of Oxford}, \orgaddress{\city{Oxford}, 
\country{England}}}

\affil[3]{\orgdiv{LIACS}, \orgname{Leiden University}, 
\orgaddress{ 
\city{Leiden}, 
\country{Netherlands}}}

\affil[4]{\orgdiv{Video \& Image Sense Lab}, \orgname{University of Amsterdam}, \orgaddress{\city{Amsterdam}, 
\country{Netherlands}}}



\abstract{
Continued advances in self-supervised learning have led to significant progress in video representation learning, offering a scalable alternative to supervised approaches by eliminating the need for manual annotations. Despite strong performance on standard action recognition benchmarks, existing video self-supervised learning methods are predominantly evaluated within narrow protocols—typically pre-training on Kinetics-400 and finetuning on similar datasets—limiting our understanding of their generalization capabilities in real-world settings.
In this work, we present a comprehensive evaluation of modern video self-supervised learning models, focusing on generalization across four key downstream factors: domain shift, sample efficiency, action granularity, and task diversity. Building on our prior work analyzing benchmark sensitivity in CNN-based contrastive learning, we extend the study to cover current state-of-the-art transformer-based video-only and video-text representation models. Specifically, we benchmark 12 transformer-based methods (7 video-only, 5 video-text) and compare them against 10 CNN-based methods, resulting in over 1100 experiments across 8 datasets and 7 downstream tasks.
 Our analysis reveals that, despite architectural advancements, transformer-based models remain sensitive to downstream conditions. No single method generalizes consistently across all factors; for instance, video-only transformers are more robust to domain shift, CNN-based models perform better on tasks requiring fine-grained temporal reasoning, and video-text transformers underperform both in several downstream settings despite large-scale pre-training. We also observe that recent transformer-based approaches do not universally outperform earlier methods. These findings provide a detailed understanding of the capabilities and limitations of current video self-supervised learning approaches and establish an extended benchmark for evaluating generalization in video representation learning. Our benchmark offers a unified protocol for future research aimed at developing robust, transferable video models. 
 Code will be made available at \href{https://github.com/fmthoker/SEVERE-BENCHMARK-plus-plus}{https://github.com/fmthoker/SEVERE-BENCHMARK-plus-plus}. 
 }
\maketitle
\section{Introduction}\label{sec1}
Video self-supervised learning has progressed at a tremendous pace in recent years, 
\eg,~\cite{ctp-wang2021unsupervised,mmvssl3-Afouras20b,qian2021spatiotemporal,piergiovanni2020evolving,rspnet-chen2020RSPNet,gdt-patrick2020multimodal,tong2022videomae,thoker2023tubelet,tong2022videomae,fan2023mgm,mvd_wang,huang2023mgmae,mme_sun,hwangeverest,salehi2025sigma,v-jepa}, offering a crucial starting point for learning rich, generalizable video representations without relying on manual annotations. This is especially important in the context of video understanding, where large-scale datasets are essential for training, but annotating such datasets is both costly and time-consuming.  At the same time, vast amounts of unlabeled video data are readily available online, presenting an untapped resource for learning meaningful representations.

Beyond cost and scalability concerns, human annotations are susceptible to errors and inconsistencies, especially in tasks involving subtle motion, ambiguous actions, or domain-specific activities. These challenges introduce annotator bias~\cite{haliburton2024uncovering,lampert2016empirical} and reduce label quality, limiting the effectiveness of supervised learning. In contrast, self-supervised learning (SSL) leverages the inherent structure and redundancy in videos --- such as temporal continuity, cross-modal correlations, and motion cues --- to define pretext tasks that guide representation learning without explicit labels.

Learning strong video representations without labels is vital in real-world applications where annotated data is scarce or hard to obtain, such as in medical videos, surveillance footage, scientific research domains, or niche industries. Furthermore, downstream tasks may involve new or rare action categories, unusual environments, or non-standard viewpoints that are not well-represented in existing labeled datasets. In these cases, SSL enables models to learn transferable and robust features that can generalize across tasks and domains. Ultimately, self-supervised learning unlocks the potential to scale video understanding to massive, diverse datasets while reducing reliance on costly annotations, making it an essential paradigm for advancing the field.

Despite rapid advancements in video self-supervised learning, the majority of the methods, \eg~\cite{clip-order-xu2019self,frame-order-misra2016shuffle,avid-cma-morgado2021audio,selavi-asano2020labelling,videomoco-pan2021videomoco,tong2022videomae,fan2023mgm,mme_sun,hwangeverest,huang2023mgmae}, remain evaluated with a narrow scope. The standard benchmarking protocol for many years~\cite{clip-order-xu2019self,frame-order-misra2016shuffle,avid-cma-morgado2021audio,selavi-asano2020labelling,videomoco-pan2021videomoco} was pre-training on the unlabeled Kinetics-400 dataset~\cite{Kinetics-400-arxiv} and evaluating by finetuning on UCF-101~\cite{UCF-101-arxiv} and HMDB-51~\cite{HMDB-51-ICCV}. More recently~\cite{tong2022videomae,fan2023mgm,mme_sun,hwangeverest,huang2023mgmae}, the evaluation has changed to finetuning on Kinetics-400~\cite{Kinetics-400-arxiv} and Something-Something-v2~\cite{SS-v2-arxiv}. %
Although these benchmarks have driven significant advancements in video self-supervised learning, they provide a limited perspective on generalizability. The pre-training and downstream datasets share similar visual characteristics and action types, making it unclear how well these methods transfer to more diverse or challenging real-world scenarios.

The narrow scope of current benchmarks poses a critical limitation: it fails to reflect the diverse real-world conditions where robust video representations are most needed, such as domain-shift, egocentric perspectives, scenarios with limited labeled finetuning data, fine-grained interactions, or complex tasks beyond standard classification. Recognizing this gap, some works have begun to evaluate on additional datasets such as ~\cite{dave2021tclr,wang2021removing,thoker2023tubelet,fan2023mgm} on Diving-48~\cite{diving}, ~\cite{xiao2021modist,yang2020video,large-scale-feichtenhofer2021large,tong2022videomae,mvd_wang} on AVA~\cite{AVA-Gu_2018_CVPR} and ~\cite{yang2020video} on EPIC-Kitchens-100~\cite{EPIC-100-arxiv}. However, these evaluations are often ad hoc, typically involving only one additional dataset and frequently without standardized baselines or comparisons to other approaches.

Some works have tried to address this problem for image self-supervised learning by performing benchmarking studies~\cite{when_contrast_work,trasnferability,contrasting_contrastive,edinburgh-ericsson2021well, goyal2019scaling, yang2020transfer, kolesnikov2019revisiting, zhai2019large, asano2019critical, newell2020useful, sariyildiz2021concept, van2021benchmarking, ericsson2021self} to investigate model transferability~\cite{trasnferability,edinburgh-ericsson2021well,newell2020useful,image-eval5-wallace2020extending} or the importance of factors like pre-training dataset~\cite{when_contrast_work,contrasting_contrastive,goyal2019scaling} and backbone architecture~\cite{kolesnikov2019revisiting}. 
Unfortunately, lessons from these works do not directly transfer to video self-supervised learning. First, video self-supervised tasks are distinct from those of images as they are designed to understand the temporal dimension of video~\cite{rspnet-chen2020RSPNet,dave2021tclr,ctp-wang2021unsupervised,yang2020video} in addition to the spatial understanding needed in images~\cite{simclr-pmlr-v119-chen20j}. Second, video is multi-modal and several methods~\cite{gdt-patrick2020multimodal,selavi-asano2020labelling,avid-cma-morgado2021audio} are designed to exploit cross or multi-modal understanding, which is again absent in image-based methods. 
We take inspiration from the benchmarking works in image self-supervised learning and perform a much-needed study to understand the generalizability of video self-supervised methods to different downstream factors. 

In the conference version of this work~\cite{thoker2022severe}, we addressed the essential need to gauge the sensitivity of existing video self-supervised methods to the current benchmark by thoroughly evaluating their performance for generalization across diverse downstream settings. In particular, we identified the problem of benchmark-sensitivity in video self-supervised learning and examined this sensitivity along the factors of domain, samples, action granularity, and task. From our extensive experiments on 9 CNN-based video self-supervised learning methods, we found that standard benchmarks in video self-supervised learning did not indicate generalization along the sensitivity factors, and vanilla supervised pre-training outperforms self-supervised pre-training.  Finally, we proposed a subset of our experiments as the \textit{SEVERE-benchmark} to evaluate the generalization capability of self-supervised learning methods.

While our prior study~\cite{thoker2022severe} offered a detailed analysis of generalization in video self-supervised learning, it was limited to CNN-based methods~\cite{moco_v2,videomoco-pan2021videomoco,selavi-asano2020labelling,pretext-contrast-DBLP:journals/corr/abs-2010-15464,avid-cma-morgado2021audio,rspnet-chen2020RSPNet,dave2021tclr,gdt-patrick2020multimodal}, the state-of-the-art video SSL methods at that time, which primarily used contrastive learning paradigms. Since then, the field has shifted dramatically toward transformer-based architectures, with two dominant trends: video-only transformers trained using masked autoencoding~\cite{tong2022videomae,fan2023mgm,mvd_wang,huang2023mgmae,mme_sun,hwangeverest,salehi2025sigma}, and video-text transformers trained on large-scale video-caption pairs using cross-modal objectives\cite{li2023unmasked,doughty2024locomotion,cheng2022vindlu,wang2023internvid}. These models now lead the state of the art and are widely used as general-purpose video encoders. However, despite their success on standard benchmarks, mainly action recognition on Kinetics-400~\cite{Kinetics-400-arxiv} and Something-Something V2~\cite{SS-v2-arxiv}, it remains unclear whether they overcome the generalization challenges previously observed or if they too are sensitive to downstream domain shifts, action granularities, task variation, and sample size.

In this work, we extend our analysis to this new generation of video representation learning models, aiming to systematically assess whether transformer-based video-only and video-text representation learning methods face the same issues and how well they generalize across diverse downstream settings. To further expand the task dimension, we incorporate Temporal Action Localization (TAL) on ActivityNet~\cite{caba2015activitynet} into our evaluation. We perform an extensive evaluation that adds over 600 new experiments with 12 transformer-based video or video-text SSL methods across 8 video datasets and 7 video understanding tasks. We observe that, similar to CNN methods, both video-only and video-text transformers show sensitivity to the evaluated downstream factors. Different methods shine for different downstream settings (\eg, for domain shift or fewer samples), with no single method showing a very strong generalization to all downstream factors. We also find that recent transformer-based methods are not necessarily the best. While video-only transformer models outperform CNN models for domain shift,  CNN-based methods are still leading for action granularities. Finally, we evaluate all the models on subsets of the original \textit{SEVERE-benchmark} and show a holistic comparison of current video representation learning methods. We hope the future works can use the updated benchmark dubbed as \textbf{SEVERE-benchmark++} to evaluate the generalization capability of all types of video representation learning methods. 

To summarize, we make the following contributions:
\begin{itemize}
\item We evaluate 7 new transformer-based video-only self-supervised learning methods across all 4 downstream sensitivity factors introduced in \cite{thoker2022severe}.
\item We evaluate 5 transformer-based video-text representation learning methods across 4 downstream sensitivity factors from \cite{thoker2022severe}.
\item We extend the task-shift downstream factor to focus on more diverse downstream tasks by including  Temporal Action Localization on ActivityNet.  
\item Overall, we conduct over 1100 experiments with 10 CNN-based video SSL methods, 7 video-only transformer SSL methods, and 5 video-text representation learning methods on 8 video datasets for 7 different video downstream tasks to demonstrate sensitivity in benchmarking video representation learning.
\item We analyse the results within individual model categories 
 (\eg~with CNNs) as well as across model categories (CNN vs. Transformers) to have a more detailed view of video representation learning capability and the current state of video SSL. 
\end{itemize}

\begin{figure*}[t]
    \centering
    \captionsetup{font=small,skip=1mm}
    \includegraphics[width=\linewidth]{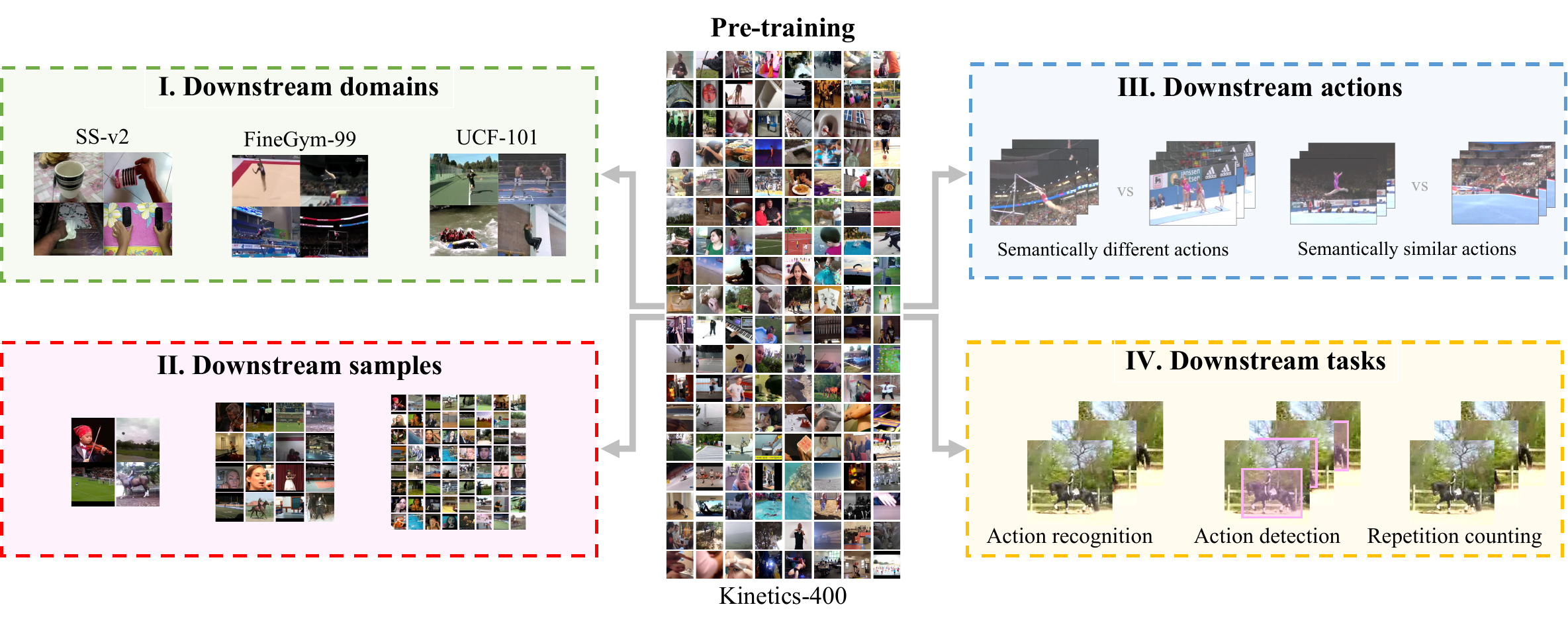}
    \caption{\textbf{Benchmark-sensitivity.} We evaluate the sensitivity of 10 CNN-based video SSL methods, 7 transformer-based video-only SSL and 5 transformer-based video-text pre-training methods for 4 downstream factors. The downstream factors vary from the pre-training source in: the domain, the samples, the actions and the task.}
    \label{fig:concept-figure}
\end{figure*}
\section{Identifying Benchmark Sensitivity}\label{sec2}
Early works in video self-supervised learning, \ie, CNN-based, evaluate their approach by pre-training on Kinetics-400~\cite{Kinetics-400-arxiv} and finetuning the learned representations for action recognition on UCF-101\cite{UCF-101-arxiv} and HMDB-51\cite{HMDB-51-ICCV}. Some works~\cite{gdt-patrick2020multimodal, dave2021tclr, pretext-contrast-DBLP:journals/corr/abs-2010-15464, ctp-wang2021unsupervised, rspnet-chen2020RSPNet, selavi-asano2020labelling,gavrilyuk2021motion, lin2021self,huang2021ascnet} also report performance on video retrieval for UCF-101 and HMDB-51 while  others~\cite{qian2021spatiotemporal,yang2020video,recasens2021broaden} also compare linear evaluation performance on Kinetics-400.
Most recent works \ie, transformer-based,  use action recognition on Kinetics-400~\cite{Kinetics-400-arxiv} and Something-Something v2~\cite{SS-v2-arxiv} as the main benchmark, and sometimes report additional results on non-standard datasets like AVA~\cite{AVA-Gu_2018_CVPR} or Diving-48~\cite{diving}. 

The problem with these evaluation setups is that the downstream dataset is either the same or shares many similarities with the pre-training dataset. For example, videos in UCF101, HMDB, and Kinetics-400 datasets are all collected from YouTube and are mostly recorded with a single camera containing a single well-positioned human actor.  
Similarly, in terms of class labels, these datasets focus on similar, coarse-grained, and mutually exclusive actions with many actions common between pre-training and downstream datasets. Besides all these data similarities, the existing evaluations also ignore a major benefit of self-supervised representation learning for videos, \ie, finetuning the representation with only a small amount of data samples and transferring to other video understanding tasks beyond action recognition. Hence, we believe the current benchmark standard is insufficiently equipped to gain a true understanding of where video self-supervised models are successful, as it cannot show the generalizability or the sensitivity of methods to factors such as domain shift, amount of finetuning data samples, action similarity or task shift. 
In this study, we identify the sensitivity of existing evaluations and thoroughly benchmark self-supervised video learning methods along four sensitivity factors as depicted in \cref{fig:concept-figure}.
\begin{enumerate}[label=\Roman*.]
\item \textbf{Downstream domain.}
First, we analyse whether features learned by self-supervised models transfer to datasets that vary in domain with respect to the pre-training dataset. 
\item \textbf{Downstream samples.} Second, we evaluate the sensitivity of self-supervised methods to the number of downstream samples available for finetuning. 
\item \textbf{Downstream actions.}
Third, we investigate if self-supervised methods learn fine-grained features required to recognize semantically similar actions. 
\item \textbf{Downstream task.}
Finally, we study the sensitivity of video self-supervised methods to the downstream task and question whether self-supervised features can be used beyond action recognition. 
\end{enumerate}

\begin{figure*}[t]
    \centering
    \includegraphics[width=\linewidth]{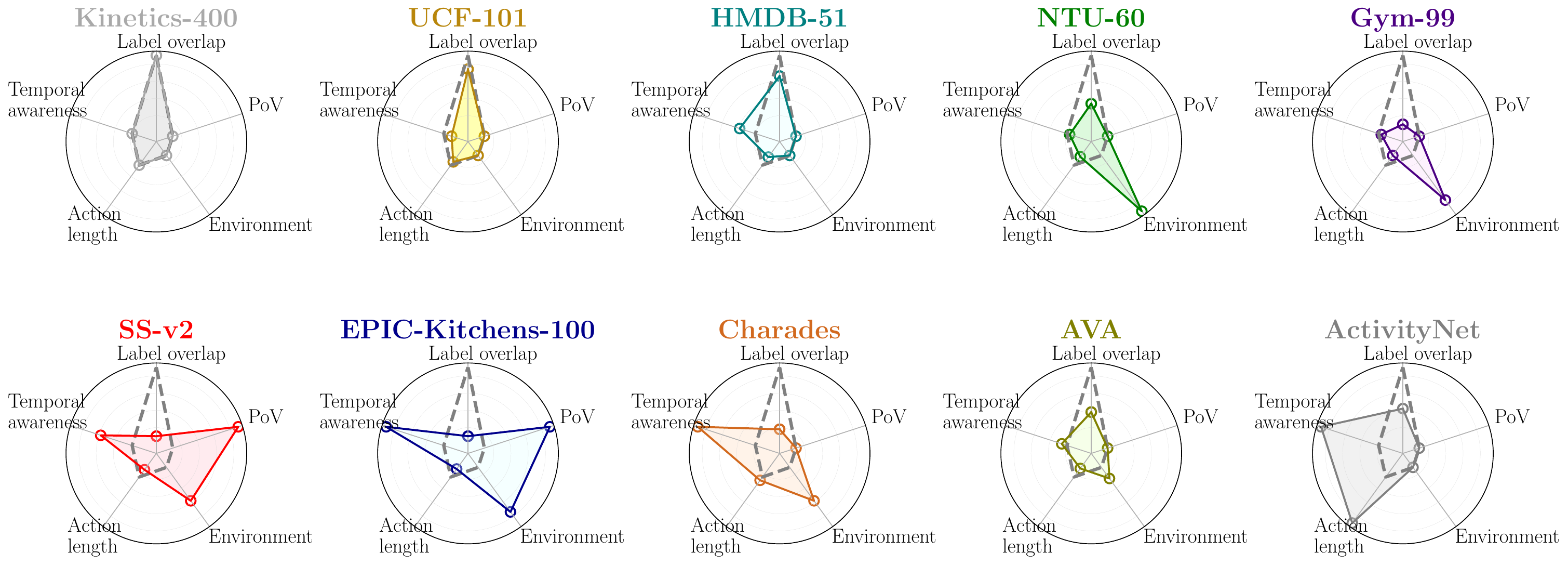}
    \caption{\small \textbf{Video dataset characteristics.}
    Characterizing domain shift in datasets via difference in label overlap, point-of-view (PoV), environment, action length and temporal awareness with Kinetics-400 (shown by dotted line). Kinetics-400 and UCF-101 are highly similar to each other, while datasets like Something-Something-v2, EPIC-Kitchens-100 and Charades have different attributes compared to Kinetics-400.}
    \label{fig:radar}
\end{figure*}

\subsection{Downstream Video Datasets}
\label{subsec:domain-shift}
We evaluate various self-supervised models along our four sensitivity factors on 8 video datasets: \textbf{UCF-101}~\cite{UCF-101-arxiv}, \textbf{NTU-60}~\cite{NTU-60-arxiv}, \textbf{FineGym} (Gym-99) \cite{Gym-99-arxiv}, \textbf{SomethingSomething-v2} (SS-v2)~\cite{SS-v2-arxiv}, \textbf{EPIC-Kitchens-100} (EK-100)~\cite{EPIC-100-arxiv}, \textbf{Charades}~\cite{charades-sigurdsson:hal-01418216} and \textbf{AVA}~\cite{AVA-Gu_2018_CVPR}, \textbf{ActivityNet}~\cite{caba2015activitynet}. They include a considerable variety in video domain, the actions they contain and cover a range of video understanding tasks. To get a sense of the differences between these downstream datasets and the Kinetics-400 source dataset, we summarize their similarity to Kinetics-400 by radar plots in \cref{fig:radar} based on several attributes. 
    \textit{Label overlap} is the fraction of actions in a target dataset that are also present in 
    Kinetics-400. We quantify this by
matching identical actions as well as manually checking for reworded versions
of the same action class.
    \textit{Point-of-view (PoV)} 
    is whether a video is recorded from a first-person (plotted as 1) or third-person viewpoint (plotted as 5).
    \textit{Environment} refers to the variety of settings contained in the dataset. Datasets are qualitatively ordered in descending order of variability with the most variable datasets plotted at the center and the least at the edge.
        \textit{Action length} is the average temporal length of the actions in seconds. 
    \textit{Temporal awareness} defines the extent to which temporal context is required to recognize or detect actions. We quantify this as the minimum number of frames required to best recognize an action. Details are in the Appendix.

\subsection{Evaluated Video Representation Learning Methods}
\label{sec:evaluated_models}

The aim of video self-supervised learning is to learn video representations from unlabeled video data. Many learning paradigms have emerged to achieve this goal from transformation prediction~\cite{jigaw1-ahsan2019video,jigsaw2-huo2021selfsupervised, jigsaw3-kim2019self,frame-order-misra2016shuffle, shuffle1-fernando2017self, shuffle2-suzuki2018learning, clip-order-xu2019self,yao2021seco} to contrastive learning~\cite{videomoco-pan2021videomoco,dave2021tclr,han2019video, yang2020video,qian2021spatiotemporal,lin2021self,diba2021vi2clr, sun2021composable,gdt-patrick2020multimodal,avid-cma-morgado2021audio,coclr, tao2020self,ma2021active,korbar2018cooperative,fmthoker_acmmm} to masked video modeling~\cite{tong2022videomae,mme_sun,fan2023mgm,yang2022motionmae,salehi2025sigma,mvd_wang,hwangeverest,huang2023mgmae}. 
Moreover,  video-text modeling has also been explored as a strong video representation learning strategy that relies on weakly supervised video-text pairs for learning.
In this work, we evaluate a diverse set of video representation learning methods from CNN and transformer-based self-supervised methods to video-language methods.
In particular, we divide the video representation learning methods into three categories as:

\subsubsection{CNN Methods}

This category consists of video SSL methods that are designed to train 3DCNNs.  The learning objective of such methods mainly involves solving different pretext tasks and contrastive learning. Pretext task methods use predictive tasks such as solving spatio-temporal jigsaw puzzles~\cite{jigaw1-ahsan2019video,jigsaw2-huo2021selfsupervised, jigsaw3-kim2019self}, rotation prediction~\cite{rotate1-jing2019selfsupervised}, 
frame and clip order~\cite{frame-order-misra2016shuffle, shuffle1-fernando2017self, shuffle2-suzuki2018learning, clip-order-xu2019self,yao2021seco}, video speed~\cite{relative-speed1-benaim2020speednet, relative-speed2-jenni2020video, playback1-cho2020self, playback2-yao2020video, playback3-wang2020self}, video  completion~\cite{vcp}, predicting motion statistics~\cite{wang2019self},  tracking random patches in video frames~\cite{ctp-wang2021unsupervised} or audio-visual clustering~\cite{multimodal-clustering1-chen2021multimodal, multimodal-clustering2-hu2019deep,  selavi-asano2020labelling,alwassel2020self}.
Contrastive learning methods 
discriminate between `positive' and `negative' pairs to learn invariances to certain data augmentations and instances either from visual-only input~\cite{videomoco-pan2021videomoco,dave2021tclr,han2019video, yang2020video,qian2021spatiotemporal,lin2021self,diba2021vi2clr, sun2021composable} or multi-modal data~\cite{gdt-patrick2020multimodal,avid-cma-morgado2021audio,coclr, tao2020self,ma2021active,korbar2018cooperative,fmthoker_acmmm}. Some methods also combine pretext and contrastive approaches~\cite{pretext-contrast-DBLP:journals/corr/abs-2010-15464, rspnet-chen2020RSPNet,pretext-contrast-2-zhang2021contrastive,taco-bai2020can, diba2021vi2clr, huang2021ascnet}. 

We consider video-based self-supervised models for CNN-based methods that achieve good performance on current benchmarks and cover a range of self-supervised paradigms in the video domain, including contrastive learning, pretext-tasks, their combination and cross-modal audio-video learning. We filter such methods based on the public availability of checkpoints, a common pre-training dataset, and a common 3D-CNN backbone. To that end, we select the  following 10 works that provide publicly available weights for an R(2+1)D-18 network~\cite{tran2018closer} pre-trained on Kinetics-400~\cite{Kinetics-400-arxiv} dataset:

\noindent\textbf{MoCo~\cite{moco_v2}} is a contrastive learning method proposed for images. Positives are different spatial augmentations of a video, while negatives are other videos. To obtain negatives beyond the current batch, MoCo maintains a queue of momentum-updated samples from previous batches.

\noindent\textbf{SeLaVi~\cite{selavi-asano2020labelling}} views the audio and visual modalities as different video augmentations and learns with cross-modal clustering.

\noindent\textbf{VideoMoCo~\cite{videomoco-pan2021videomoco}} extends MoCo to video by learning invariance to temporal augmentations as well as spatial. It uses adversarial dropout to remove the frames most important to the model prediction. 

\noindent\textbf{Pretext-Contrast~\cite{pretext-contrast-DBLP:journals/corr/abs-2010-15464}} combines pretext tasks with contrastive learning. As its pretext task, it uses video cloze procedure~\cite{vcp} which predicts the augmentations applied to a video. For contrastive learning, it uses different temporal shifts to form different video clips.

\noindent\textbf{RSPNet~\cite{rspnet-chen2020RSPNet}} also combines pretext and contrastive tasks, with a focus on video speed. The pretext task predicts the relative difference in speed between two versions of a video, while the contrastive task creates positives and negatives with speed-based augmentations.

\noindent\textbf{AVID-CMA~\cite{avid-cma-morgado2021audio}} is an audio-visual contrastive learning method. It first uses cross-modal contrastive learning followed by within modality contrastive learning, where additional positives with high audio and visual similarity are sampled.

\noindent\textbf{CtP~\cite{ctp-wang2021unsupervised}} performs self-supervised learning through a ``catch the patch'' pretext task. The goal is to predict the trajectory of an image patch, which is moved through the video sequence.

\noindent\textbf{TCLR~\cite{dave2021tclr}} is a contrastive method that encourages features to be distinct across the temporal dimension by using clips from the same video as negatives rather than positives. 
It also uses extensive spatial augmentations.

\noindent\textbf{GDT~\cite{gdt-patrick2020multimodal}} is a multi-modal contrastive method that 
encourages invariance to spatial augmentations, audio and visual modalities and temporal 
reversal, while encouraging variance to different temporal shifts.

\noindent\textbf{Tubelet-Contrast~\cite{thoker2023tubelet}} is a contrastive learning method designed to learn motion-centric video representations. It maximizing feature similarity among videos sharing identical synthetic motions while minimizing similarity between those with differing synthetic motions.

\noindent\textbf{Supervised}. We also compare with supervised pre-training and include an R(2+1)D-18 model trained on Kinetics-400 for action recognition with ground-truth labels. 

\subsubsection{Video-only Transformer Methods} In recent years, transformer-based methods have become a de facto choice for learning powerful self-supvervised video representations with state-of-the-art  performance. Following the success of masked auto encoders MAE~\cite{he2022masked} for image representation learning, many works have extended the idea to learn video representations via a mask and predict learning strategy~\cite{tong2022videomae, fan2023mgm, mme_sun, huang2023mgmae, salehi2025sigma, yang2022motionmae}.

We consider the top video SSL transformer-based methods 
and filter them based on the public availability of checkpoints for a ViT-B~\cite{dosovitskiy2020image} network trained on Kinetics-400~\cite{Kinetics-400-arxiv}. We select the following 7 works:

\noindent\textbf{VideoMAE}~\cite{tong2022videomae} is the extension of MAE~\cite{he2022masked} for video. The task is to reconstruct masked tubes of the video (\ie~patches across time) which are randomly masked with a 90\% ratio.

\noindent\textbf{MGMAE}~\cite{huang2023mgmae} utilizes motion-guided masking, leveraging optical flow to align masking maps across adjacent frames, ensuring consistent masking volumes over time.

\noindent\textbf{MGM}~\cite{fan2023mgm} also replaces random masking in VideoMAE with motion-guided masking. Here, the motion-guide leverages motion vectors from the H.264 video codec. 

\noindent\textbf{EVEREST}~\cite{hwangeverest} uses masked video pre-training on only motion-rich tokens, discarding uninformative ones.  

\noindent \textbf{MVD}~\cite{mvd_wang} uses a masked feature distillation framework.
The feature representations of the image MAE and the video MAE teacher are utilized as reconstruction targets for masked video feature modeling. 

\noindent\textbf{SIGMA}~\cite{salehi2025sigma} reconstructs semantic features instead of pixels and replaces the MSE loss with a Sinkhorn-Guided clustering loss. 
 
\noindent\textbf{MME}~\cite{mme_sun} 
diverges from predicting appearance content to predicting motion trajectories that capture future position and shape changes in the mask-and-predict task. 

\noindent\textbf{Supervised} As for CNNs, we also include a supervised ViT-B model that is trained for action recognition on Kinetics-400 with ground-truth labels.
 
\subsubsection{Video-text Transformers}  Recently, vision-language modeling~\cite{wang2023internvid,zhao2024videoprism} has emerged as a powerful paradigm for learning strong visual representations  from paired video and text modality. Thus, we also include video-text modeling methods in this exploration to analyze their performance on diverse downstream tasks and to compare with video-only pre-training.
Similar to image-text methods e.g. CLIP~\cite{radford2021learning},  the learning objective of such methods is to align the outputs of a video and a text encoder in a cross-modal contrastive manner. After pre-training, we only use the video encoder for various downstream tasks, without text input.

Different video-language methods use different video-text datasets for pre-training without any common standard. Thus, we choose methods with diverse pre-training datasets and filter based on the availability of the ViT-B checkpoint for the video encoder. We select the following 5 methods:

\noindent\textbf{CLIP}~\cite{radford2021learning} learns to align images and text by jointly embedding them into a shared feature space using contrastive learning. While image-based, it has been adapted for various video tasks such as action recognition~\cite{wang2021actionclip,ni2022expanding,rasheed2023fine}. The model is trained on 400 million image-text pairs.

\noindent\textbf{ViCLIP}~\cite{wang2023internvid} starts from CLIP initialized weights to align video and the text modalities with a second stage cross-modal pre-training on the 10 million video-text pairs in InternVid~\cite{wang2023internvid}.

\noindent\textbf{VindLU}~\cite{cheng2022vindlu} trains with three distinct learning objectives: video-text contrastive learning, video-text matching and masked language modeling. The model uses 25 million video-text pairs~\cite{bain2021frozen} for pre-training.

\noindent\textbf{LocoMotion}~\cite{doughty2024locomotion}  learns motion-focused video-language representations. To form the training data, synthetic motions are overlaid on videos. The parameters of this synthetic motion are then used to generate corresponding captions that describe object movements and their temporal progression. The model is trained with the same losses as VindLU\cite{cheng2022vindlu} and uses generates synthetic video-text pairs with 2.5 million videos from WebVid~\cite{bain2021frozen}.

\noindent\textbf{UMT}~\cite{li2023unmasked} is a two-stage pre-training method that leverages video-only and video-text data for representation learning. The first stage trains a VideoMAE-based framework using video-only data from Kinetics-700. The second stage further trains this representation with the same learning objectives and data as VindLU. 

Since we use publicly available models we cannot control the exact pre-training setup of each method. There are differences in the training regime for different methods, such as the number of epochs, data augmentations, batch size, etc. Details of these differences are provided in the appendix. However, as mentioned above, all methods in the same category  use the common backbone network \ie R2+1D-18 for CNNs and ViT-B for video-only  or video-text transformers.  Thus for each category,  we can evaluate their downstream abilities in exactly the same way. 
To finetune for downstream tasks we attach a task-dependent head at the last layer of the pre-trained video backbone to produce label predictions for the corresponding task. For a fair comparison, during training in the downstream tasks, we use the same setup for all the models within the same category.
\section{Sensitivity Factor I: Downstream Domain}\label{sec:factor_1}
\medskip
%

We first investigate to what extent video representation learning methods learn features that are applicable to action recognition in any domain. 
We evaluate the suite of pre-trained models on Kinetics-400 (K-400)~\cite{Kinetics-400-arxiv}, UCF-101~\cite{UCF-101-arxiv}, NTU-60~\cite{NTU-60-arxiv}, Gym-99~\cite{Gym-99-arxiv}, SS-v2~\cite{SS-v2-arxiv} and EK-100~\cite{EPIC-100-arxiv} for the task of action recognition.
It is worth noting that besides their variety in the domain, these datasets demonstrate varieties in the number of training data (9.5k - 168k samples) and the number of categories (60 - 300 classes). We attach a single classification layer to the pre-trained backbone and evaluate the model's performance on the downstream task in two settings. First, \textbf{finetuning} where both the pre-trained backbone and the classification layer are trained for action classification. Second, \textbf{linear evaluation} that only trains the classification layer using the frozen features from the pre-trained backbones. 
We follow the standard splits proposed in the original datasets and report video-level top-1 accuracy on the test sets. The details about splits, pre-processing, and training for each dataset are provided in the appendix.

\subsection{CNNs}
\medskip
\noindent\textbf{Finetuning.} The left part of \cref{domain_shift_cnns} shows the results of finetuning for CNN-based models. All the methods share the same architecture \ie~R(2+1)D-18~\cite{tran2018closer}.
From the results, it is clear that all self-supervised methods are very effective for in-domain finetuning on Kinetics-400 (K-400) and UCF-101, with a significant gap between training from scratch and all self-supervised methods. This gap is reduced as the difference between the pre-training dataset Kinetics-400 and the downstream domain increases (see \cref{fig:radar}).  SeLaVi, MoCo and AVID-CMA in particular, are evidence of this as these methods suffer when datasets have higher temporal awareness and less label overlap with Kinetics-400. When moving from UCF-101 to NTU-60 and Gym-99 there is a change in the ordering of self-supervised methods. This demonstrates a high performance on Kinetics-400 or UCF-101 does not guarantee that a self-supervised model is generalizable to other domains. Similarly a lower performance on Kinetics-400 doesn't indicate the model will not generalize across domains \eg~VideoMoCo and Tubelet-Contrast are among the top performers on NTU-60, SS-v2, and EK-100 while being among the bottom ones for Kinetics-400. 

The change in ranking is even more prominent for SS-v2 and EK-100, which require the most temporal awareness and also shift to a first-person viewpoint. This is particularly noticeable for AVID-CMA. On these datasets, MoCo has similar results to no pre-training, which is evidence that video-specific self-supervised learning methods are needed and that image-based methods are insufficient.
In addition, supervised pre-training achieves good performance across the board, outperforming self-supervised methods on the most similar domain (UCF-101) as well as the most dissimilar domains (SS-v2 and EK-100). Amidst the models tested, Tubelet-Contrast, CtP, RSPNet, and TCLR stand out as the self-supervised pre-training methods most generalizable to different domains. CtP and Tubelet-Contrast explicitly aim to learn motion-focused video representations and demonstrate the best generalization highlighting the impact of motion awareness for learning generalizable features.

\noindent\textbf{Linear Evaluation.} The right part of \cref{domain_shift_cnns} shows the results for linear classification for CNN-based models.
As with finetuning, the ranking among the self-supervised methods changes as the domain difference between the pre-training and the downstream dataset increases. For example, VideoMoCo ranks lower than GDT and RSPNet for UCF-101 and Kinetics-400 but ranks higher than both for all other datasets. This again demonstrates that performance on UCF-101 and Kinetics-400, which are used as standard benchmarks by most CNN-based methods~\cite{qian2021spatiotemporal, recasens2021broaden, yang2020video}, does not give a complete picture of a self-supervised model's success.
For UCF-101 and Kinetics-400, most contrastive self-supervised models learn highly discriminative features compared to the non-contrastive models. This can be seen by comparing contrastive models AVID-CMA, GDT and RSPNet to non-contrastive SeLaVi and CtP. %
Interestingly, Tubelet-Contrast and CtP which are the best-performing methods for finetuning achieve the worst results on linear evaluation likely due to use of synthetic motions in the pre-training. While this provides model which can be more easily adapting in finetuning, it means the frozen feature space is worse at distinguishing real videos. 
This also demonstrates the lack of any correlation between finetuning and linear evaluation performance.

From the NTU-60 and Gym-99 results we observe that as the label overlap between the pre-training and the downstream dataset decreases, the performance gap between finetuning and linear evaluation increases considerably. 
This is true for both supervised and self-supervised pre-training. The most generalizable methods in the linear classification setting are contrastive methods VideoMoCo and AVID-CMA as well as supervised pre-training. Interestingly, there are cases where VideoMoCo and AVID-CMA even outperform supervised pre-training, namely for NTU-60, Gym-99 and SS-v2. 

\begin{table*}[t]
\captionsetup{font=small,skip=2mm}
        \caption[]{\textbf{Factor I: Sensitivity to Downstream Domain for CNN Methods.} Video CNN-based self-supervised methods evaluated across datasets with increasing domain shift with respect to the source dataset (see \cref{fig:radar}).  
        Colors denote relative rankings across methods for each dataset, ranging from \textcolor{lowcolor}{low} \begin{tikzpicture}%
        \pgfplotscolorbardrawstandalone[%
        colormap name=PiYG,%
        colorbar horizontal,%
        colorbar style={%
            height=0.18cm,%
            width=2cm,%
            hide axis,%
            }%
        ]%
    \end{tikzpicture} \textcolor{highcolor}{high}. The ranking of CNN-based methods is very sensitive to the downstream domain for both finetuning and linear classification. As the domain shift increases, the ranking becomes less and less correlated with the standard video SSL benchmarks \ie~UCF-101 or K400 finetuning performance.}
    \centering
    \midsepremove
    \resizebox{0.9\linewidth}{!}{\begin{tabular}{
    l\C{64.3}{68.1}\C{83.3}{91.3}\C{92.8}{94.3}\C{88.9}{92.8}\C{52.0}{60.2}\C{26.4}{43.6}c\C{7.2}{46.0}\C{37.9}{91.7}\C{15.7}{53.9}\C{20.2}{45.1}\C{4.5}{16.6}\C{18.8}{26.6}}
    \toprule
    \addlinespace[0.1cm]
    \multirow{2}{*}{\textbf{Pre-training}} & 
    \multicolumn{6}{c}{\textbf{Finetuning}} & & 
    \multicolumn{6}{c}{\textbf{Linear Evaluation}} \\
    \addlinespace[0.04cm]
    \cmidrule{2-7} \cmidrule{9-14}
      \addlinespace[0.1cm]
       & \multicolumn{1}{c}{K-400}  & \multicolumn{1}{c}{UCF-101} &  \multicolumn{1}{c}{NTU-60} & \multicolumn{1}{c}{Gym-99} & \multicolumn{1}{c}{SS-v2} & \multicolumn{1}{c}{EK-100} & & \multicolumn{1}{c}{K-400} & \multicolumn{1}{c}{UCF-101} &  \multicolumn{1}{c}{NTU-60} & \multicolumn{1}{c}{Gym-99} & \multicolumn{1}{c}{SS-v2} & \multicolumn{1}{c}{EK-100}\\
         \midrule
           \addlinespace[0.01cm]
         None                & \multicolumn{1}{c}{60.1}    & \multicolumn{1}{c}{77.3} & \multicolumn{1}{c}{92.9} & \multicolumn{1}{c}{89.8} & \multicolumn{1}{c}{57.1} & \multicolumn{1}{c}{25.7} & & \multicolumn{1}{c}{-}& \multicolumn{1}{c}{-} & \multicolumn{1}{c}{-} & \multicolumn{1}{c}{-} & \multicolumn{1}{c}{-} & \multicolumn{1}{c}{-}\\
           \addlinespace[0.01cm]
         \midrule
        \addlinespace[0.01cm]
         MoCo\cite{moco_v2}                 &64.3   & 83.3 & 93.4 & 90.7 & 57.1 & 26.4 & & 34.5 & 65.4 & 16.0 & 21.2 & 7.4 & 21.4 \\
         VideoMoCo\cite{videomoco-pan2021videomoco}            &65.0    & 84.9 & 94.1 & 90.3 & 59.0 & 43.6 && 31.0 & 66.3 & 51.6 & 41.6 & 19.5 & 25.7 \\
         SeLaVi\cite{selavi-asano2020labelling}               &65.5    & 85.2 & 92.8 & 88.9 & 56.2 & 33.8 && 24.1 & 51.2 & 15.7 & 20.2 & 4.5 & 22.4 \\
         Pretext-Contrast\cite{pretext-contrast-DBLP:journals/corr/abs-2010-15464}     &66.1    & 87.7 & 93.9 & 90.5 & 56.9 & 34.3 && 22.4 & 57.2 & 17.6 & 30.0 & 10.9 & 20.0 \\
         RSPNet\cite{rspnet-chen2020RSPNet}               &66.4    & 88.7 & 93.9 & 91.1 & 59.0 & 42.7 && 46.0 & 76.6 & 33.5 & 32.2 & 12.5 & 24.9 \\
         AVID-CMA\cite{avid-cma-morgado2021audio}              & 66.6   & 88.8 & 94.0 & 90.4 & 52.0 & 29.9 && 43.5 & 78.1 & 53.9 & 45.1 & 16.1 & 22.5 \\
         CtP\cite{ctp-wang2021unsupervised}                   &67.1   & 90.1 & 94.3 & 92.0 & 59.6 & 42.8 && 7.6 & 37.9 & 22.6 & 30.6 & 12.2 & 20.0 \\
         TCLR\cite{dave2021tclr}                 &68.1    & 90.8 & 94.1 & 91.6 & 59.8 & 36.2 && 19.9 & 63.3 & 33.5 & 33.0 & 10.8 & 21.8 \\
                  Tubelet-Contrast\cite{thoker2023tubelet}                  & 65.8    & 91.0 & 93.7 & 92.8 & 60.2 & 43.1 && 7.2 & 37.1 & 22.1 & 28.5 & 11.4 & 18.8 \\
         GDT\cite{gdt-patrick2020multimodal}                  &67.1    & 91.3 & 93.9 & 90.5 & 58.0 & 37.3 && 38.6 & 75.7 & 38.2 & 34.2 & 11.9 & 25.3 \\
        \addlinespace[0.01cm]
         \midrule
        \addlinespace[0.01cm]
         Supervised           & \multicolumn{1}{c}{-}     & \multicolumn{1}{c}{91.4} & \multicolumn{1}{c}{93.9} & \multicolumn{1}{c}{92.1} & \multicolumn{1}{c}{60.8} & \multicolumn{1}{c}{47.7} && \multicolumn{1}{c}{-} & \multicolumn{1}{c}{91.7} & \multicolumn{1}{c}{45.5} & \multicolumn{1}{c}{42.7} & \multicolumn{1}{c}{16.6} & \multicolumn{1}{c}{26.6} \\
        \addlinespace[0.01cm]
         \bottomrule
    \end{tabular}
    }

    \label{domain_shift_cnns}
\end{table*}
\begin{table*}[h]
\captionsetup{font=small,skip=2mm}
         \caption[]{\textbf{Factor I: Sensitivity to Downstream Domain for Video-Only Transformer Methods.} Video self-supervised methods evaluated across datasets with increasing domain shift with respect to the source dataset (see \cref{fig:radar}).  
         Colors denote relative rankings across methods for each dataset, ranging from \textcolor{lowcolor}{low} \begin{tikzpicture}%
      \pgfplotscolorbardrawstandalone[%
        colormap name=PiYG,%
        colorbar horizontal,%
        colorbar style={%
          height=0.18cm,%
          width=2cm,%
          hide axis,%
        }%
      ]%
    \end{tikzpicture} \textcolor{highcolor}{high}. The ranking of video-only transformer methods is less domain-sensitive for both finetuning and linear classification and is mildly correlated with performance on the current video SSL benchmarks \ie~finetuning on K400 and SSv2. }
    \centering
    \midsepremove
    \resizebox{0.9\linewidth}{!}{\begin{tabular}{
    l\C{79.0}{81.3}\C{93.3}{96.0}\C{90.0}{94.0}\C{82.5}{90.7}\C{68.0}{71.1}\C{60.2}{63.5}c\C{14.1}{47.5}\C{49.1}{80.7}\C{11.5}{34.4}\C{22.7}{30.1}\C{12.2}{21.7}\C{29.7}{34.2}}
    \toprule
    \addlinespace[0.1cm]
     \multirow{2}{*}{\textbf{Pre-training}} & \multicolumn{6}{c}{\textbf{Finetuning}} & & \multicolumn{6}{c}{\textbf{Linear Evaluation}} \\
    \addlinespace[0.04cm]
    \cmidrule{2-7} \cmidrule{9-14}
      \addlinespace[0.1cm]
       & \multicolumn{1}{c}{K-400}  & \multicolumn{1}{c}{UCF-101} &  \multicolumn{1}{c}{NTU-60} & \multicolumn{1}{c}{Gym-99} & \multicolumn{1}{c}{SS-v2} & \multicolumn{1}{c}{EK-100} & & \multicolumn{1}{c}{K-400} & \multicolumn{1}{c}{UCF-101} &  \multicolumn{1}{c}{NTU-60} & \multicolumn{1}{c}{Gym-99} & \multicolumn{1}{c}{SS-v2} & \multicolumn{1}{c}{EK-100}\\
         \midrule
           \addlinespace[0.01cm]
         None           & \multicolumn{1}{c}{69.1}         & \multicolumn{1}{c}{52.1} & \multicolumn{1}{c}{60.6}  & \multicolumn{1}{c}{50.0} & \multicolumn{1}{c}{49.8} & \multicolumn{1}{c}{35.4} && \multicolumn{1}{c}{-} & \multicolumn{1}{c}{-} & \multicolumn{1}{c}{-} & \multicolumn{1}{c}{-} & \multicolumn{1}{c}{-} & \multicolumn{1}{c}{-} \\
           \addlinespace[0.01cm]
         \midrule
        \addlinespace[0.01cm]
         EVEREST~\cite{hwangeverest} & 79.0& 93.3 & 92.3 & 88.2 & 68.0 & 62.2 && 14.1 & 51.8 & 20.3 & 23.3 & 14.5 & 30.5 \\
         MVD~\cite{mvd_wang}        & 79.7   & 94.0 & 90.0 & 82.5  & 68.5 & 60.2 && 18.7 & 49.1 & 11.5 & 22.7 & 12.2 & 29.7 \\
         MGMAE~\cite{huang2023mgmae}       & 79.9  & 95.2 & 92.9 & 87.2  & 68.9 & 63.0 && 24.9 & 64.4 & 25.3 & 26.1 & 16.8 & 33.2 \\
         VideoMAE~\cite{tong2022videomae}     & 80.0 & 94.2 & 91.1 & 86.8 & 68.6 & 62.7 && 20.7 & 58.6 & 24.3 & 23.9 & 17.5 & 33.2 \\
         MGM~\cite{fan2023mgm} & 80.6& 96.0 & 93.0 & 89.1 & 71.1 & 62.9 && 19.8 & 62.5 & 31.6 & 25.8 & 21.7 & 32.4 \\
         MME~\cite{mme_sun}    & 80.7       & 95.8 & 93.1 & 90.7  & 70.1 & 62.9 && 19.1 & 56.0 & 32.9 & 29.0 & 16.6 & 32.2 \\
         SIGMA~\cite{salehi2025sigma} & 81.3 & 95.4 & 94.0 & 89.7 & 70.9 & 63.5 && 47.5 & 80.7 & 34.4 & 30.1 & 20.8 & 34.2 \\
         \midrule
         \addlinespace[0.01cm]
         Supervised & \multicolumn{1}{c}{-} & \multicolumn{1}{c}{93.6} & \multicolumn{1}{c}{87.2} & \multicolumn{1}{c}{76.5}  & \multicolumn{1}{c}{59.5} & \multicolumn{1}{c}{56.9} && \multicolumn{1}{c}{-} & \multicolumn{1}{c}{92.4} & \multicolumn{1}{c}{60.3} & \multicolumn{1}{c}{42.0} & \multicolumn{1}{c}{24.8} & \multicolumn{1}{c}{37.5} \\
        \addlinespace[0.01cm]
         \bottomrule
    \end{tabular}
    }

    \label{domain_shift_transformer}
\end{table*}
\begin{table*}[h]
\captionsetup{font=small,skip=2mm}
         \caption[]{\textbf{Factor I: Sensitivity to Downstream Domain for Video-Text Transformer Methods.} Video-text methods evaluated across datasets with diverse domains.  
         Colors denote relative rankings across methods for each dataset, ranging from \textcolor{lowcolor}{low} \begin{tikzpicture}%
      \pgfplotscolorbardrawstandalone[%
        colormap name=PiYG,%
        colorbar horizontal,%
        colorbar style={%
          height=0.18cm,%
          width=2cm,%
          hide axis,%
        }%
      ]%
    \end{tikzpicture} \textcolor{highcolor}{high}. The ranking of methods is domain-sensitive for both finetuning and linear classification.}
    \centering
    \midsepremove
    \resizebox{\linewidth}{!}{\begin{tabular}{
    llc\C{78.2}{82.4}\C{92.0}{96.0}\C{92.0}{94.0}\C{88.0}{89.9}\C{66.7}{70.1}\C{46.3}{55.0}c\C{54.4}{65.4}\C{77.5}{88.0}\C{21.2}{43.8}\C{20.7}{31.2}\C{11.3}{18.9}\C{25.1}{28.4}}
    \toprule
    \addlinespace[0.1cm]
     \multicolumn{2}{c}{\textbf{Pre-training}} & & \multicolumn{6}{c}{\textbf{Finetuning}} & & \multicolumn{6}{c}{\textbf{Linear Evaluation}} \\
    \addlinespace[0.04cm]
    \cmidrule{1-2} \cmidrule{4-9} \cmidrule{11-16}
      \addlinespace[0.1cm]
      \multicolumn{1}{l}{Method} 
        & \multicolumn{1}{l}{Dataset} && 
        \multicolumn{1}{c} {K-400}  & \multicolumn{1}{c}{UCF-101} &  \multicolumn{1}{c}{NTU-60} & \multicolumn{1}{c}{Gym-99} & \multicolumn{1}{c}{SS-v2} & \multicolumn{1}{c}{EK-100} & & \multicolumn{1}{c}{K-400} & \multicolumn{1}{c}{UCF-101} &  \multicolumn{1}{c}{NTU-60} & \multicolumn{1}{c}{Gym-99} & \multicolumn{1}{c}{SS-v2} & \multicolumn{1}{c}{EK-100}\\
         \midrule
         None       & -   && \multicolumn{1}{c}{69.1}         & \multicolumn{1}{c}{52.1} & \multicolumn{1}{c}{60.6}  & \multicolumn{1}{c}{50.0} & \multicolumn{1}{c}{49.8} & \multicolumn{1}{c}{35.4} && \multicolumn{1}{c}{-} & \multicolumn{1}{c}{-} & \multicolumn{1}{c}{-} & \multicolumn{1}{c}{-} & \multicolumn{1}{c}{-} & \multicolumn{1}{c}{-} \\
         \midrule
           \addlinespace[0.01cm]
         LocoMotion\cite{doughty2024locomotion} & WebVid-2.5M~\cite{bain2021frozen}&& 78.2& 92.0 & 92.5 & 89.5  & 66.7 & 46.3  && 49.8 & 81.4 & 32.4 & 29.1 & 15.8 & 27.4 \\
         VindLU\cite{cheng2022vindlu} & WebVid-25M~\cite{bain2021frozen}&& 79.1& 94.5 & 92.0 & 89.3  & 66.7 & 47.2  && 54.4 & 85.4 & 43.8 & 31.2 & 17.2 & 28.4 \\
         UMT\cite{li2023unmasked} & K700\cite{carreira2019short}+WebVid-25M\cite{bain2021frozen}&& 81.7& 96.0 & 94.0 & 89.9  & 70.1 & 50.1 && 63.5 & 88.0 & 42.9 & 26.4 & 18.8 & 28.2 \\
                  CLIP~\cite{radford2021learning} & CLIP-400M~\cite{radford2021learning} && 81.8& 93.6 & 93.2 & 88.0  & 66.7 & 50.3 && 56.5 & 77.5 & 21.2 & 20.7 & 11.3 & 25.1 \\
         VICLIP~\cite{wang2023internvid} & InternVid~\cite{wang2023internvid} && 82.4& 95.2 & 93.7 & 89.7  & 67.9 & 55.0 && 65.3 & 86.7 & 35.1 & 27.3 & 18.9 & 27.3 \\
         \bottomrule
    \end{tabular}
    }

    \label{domain_shift_vision_language}
\end{table*}

\subsection{Video-only Transformers.}
\noindent\textbf{Finetuning.} The left part of \cref{domain_shift_transformer} shows the results of finetuning for transformer models that are trained in a self-supervised manner with only the video modality. All methods share the same network architecture ViT-B~\cite{dosovitskiy2020image} with joint space-time attention~\cite{arnab2021vivit,liu2022video}.
From the results, it is clear that all self-supervised methods are effective on all datasets, showing a significant advantage over training from scratch. We also observe a that self-supervised pre-training outputperforms supervised pre-training. This difference is particularly noticable for the larger domain shifts, \eg~with Gym-99 and SSv2. 

Unlike in CNNs, the change in the downstream domain does not drastically impact the rank of video-only transformer SSL models. This is particularly noticeable for datasets with significant shifts, \ie~Gym-99, SS-v2 and EK-100. While the ranking of the transformer-based methods is still different per dataset there is a strong positive correlation between Kinetics-400/UCF-101 performance and the performance on the other datasets, which was not present with the CNN-based methods.
However, we do observe that the performance of methods decrease as the domains get more different (\eg~Gym-99, SS-v2 and EK-100) highlighting the potenital for more increase in such domains than standard UCF-101 or Kinetics-400.

Among the SSL methods, SIGMA, MME, and MGM are the most generalizable across all domain shifts. This suggests that non-pixel targets (MME, SIGMA) and motion-focused masking (MME, MGM) result in learning more generalizable features for reconstruction tasks. 

Finally, we observe that transformers-based approaches are not always better than CNN ones. For the largest domain shifts EK-100 and SSv2 the benefit of transformer-based approaches is significant, however CNN-based approaches outperform transformer-based ones for NTU-60 and Gym-99.

\noindent\textbf{Linear Evaluation.} The right part of \cref{domain_shift_transformer} shows the results for linear classification for video-only transformer models. 
Different from finetuning, the self-supervised methods dramatically lag behind the supervised pre-training in linear evaluation. A reason for this could be the large difference between the pretext task of masked video reconstruction and action recognition. This means that while the model is generalizable, frozen features perform poorly. For the datasets with less domain shift, CNN-based approaches actually outperform transformer-based ones. This again hints at the pretext task being the cause, since the contrastive task used in the better performing CNN methods is much closer to the action recognition task.

Unlike finetuning, the rank of the SSL methods for different domains is less correlated with Kinetics-400 performance. Interestingly, the gap between self-supervised and supervised also decreases as the domain shift gets larger, again highlighting that transformer-based approaches are reasonably robust to domain-shift.

Among all the methods, SIGMA stands out as performing the best across almost all the datasets, potentially due to using DINO feature clustering for guidance instead of pixel reconstruction used by others.

\subsection{Video-text Transformers.}
\noindent\textbf{Finetuning.} The left part of \cref{domain_shift_vision_language} shows the results of finetuning for transformer models that pre-train with paired vision and language data. All the methods share the same network architecture ViT-B with joint space-time attention but are trained with different data as discussed in \cref{sec:evaluated_models}. Note that while the models are pre-trained with language, we only use the vision encoder for the downstream task. 

First, we observe that better performance on Kinetics is not indicative of generalization to other domains or even good performance on UCF-101. 
Amidst the methods, UMT and VICLIP stand out, the former being most generalizable to the first four datasets, and VICLIP being the best on EK-100. The strong generalization ability of UMT is likely due to the inclusion of the video-only VideoMAE pre-training stage which we see makes for domain generalizable representation from Table~\ref{domain_shift_transformer}. Interestingly, LocoMotion achieves on par performance with VinddLU while pre-training with significantly less data. 

Comparing video-text pre-training (\cref{domain_shift_vision_language}) with video-only transformers (\cref{domain_shift_transformer}), we observe that despite using significantly large amounts of data (10-25M pairs vs 240K videos), video-text underperform on datasets with a high domain variance like Gym-99, SSv2, and EK-100. While these methods are able to learn the connection between video and language it seems that this does not result in a more generalizable video representation. One of the reasons for this could be that the alignment with video and text can often be achieved by learning high-level spatial semantics without learning about more temporal semantics required for generalization. 
This is also evident from the higher performance of video-language methods on domains with a low variance like UCF101, NTU60, and K400, which do not require a lot of temporal awareness, as shown in \cref{fig:radar}.

\noindent\textbf{Linear Evaluation.} The right part of \cref{domain_shift_vision_language} shows the results for linear classification for video-text transformer models. As with finetuning, the ranking among these methods changes as the domain difference between the pre-training and the downstream dataset increases. CLIP always shows the lowest performance, while the other methods show advantages on different datasets.

Again, compared to video-only transformers (\cref{domain_shift_transformer}), vision-language models obtain a significant improvement on datasets with low domain variance (Kinetics 400, UCF-101, and NTU-60) but are on par or worse for datasets with high domain variance (Gym-99, SS-v2, EK-100).
This again validates that even with large-scale training, video-text methods are skewed towards high-level spatial semantics and do not generalize well to temporal or motion-focused domains. As with CNN and video-only transformer models, a good performance on one domain does not indicate the generalization capability for vision-language models. Thus, current video representation learning benchmarks like K400 or UCF101 or SSV2 are not well representative for evaluating video-text models.

\begin{myboxi}[]{LimeGreen!60}
\noindent\textbf{Conclusion.}
%
%
Current benchmarks, like K400, UCF-101, do not reliably reflect the generalization ability of video representation models across downstream domains. This is particularly true for CNN-based approaches and video-text transformer models, where the ranking of methods changes substantially across datasets with both full finetuning and linear classification. Finetuned video-only transformer methods are much more robust to the downstream domain, however, datasets with a bigger domain shift show more potential for improvement. 
\end{myboxi}

\section{Sensitivity Factor II: Downstream Samples}
\label{sec:factor_2}

The previous section analyzed sensitivity to the downstream domain by evaluating performance on several different datasets. However, finetuning on each of these datasets uses a large number of labeled examples, which means training from scratch can already obtain good performance. Not all domains and use cases have ample labeled video examples available, thus we investigate what the impact of the number of finetuning samples is and whether video representation learning methods can be beneficial in scenarios where we have little data to finetune with. We vary the amount of finetuning data, beginning from 1000 videos, sampled uniformly from the classes, and double the amount until we reach the full training set size. We report on four of the downstream datasets from the previous section: UCF-101, NTU-60, Gym-99 and SS-v2. The results for all three categories of evaluated models are summarized in \cref{fig:training-data-size}.


\afterpage{
\begin{figure*}[t!]
\captionsetup{font=small,skip=2mm}
    \centering
    \includegraphics[width=0.8\linewidth]{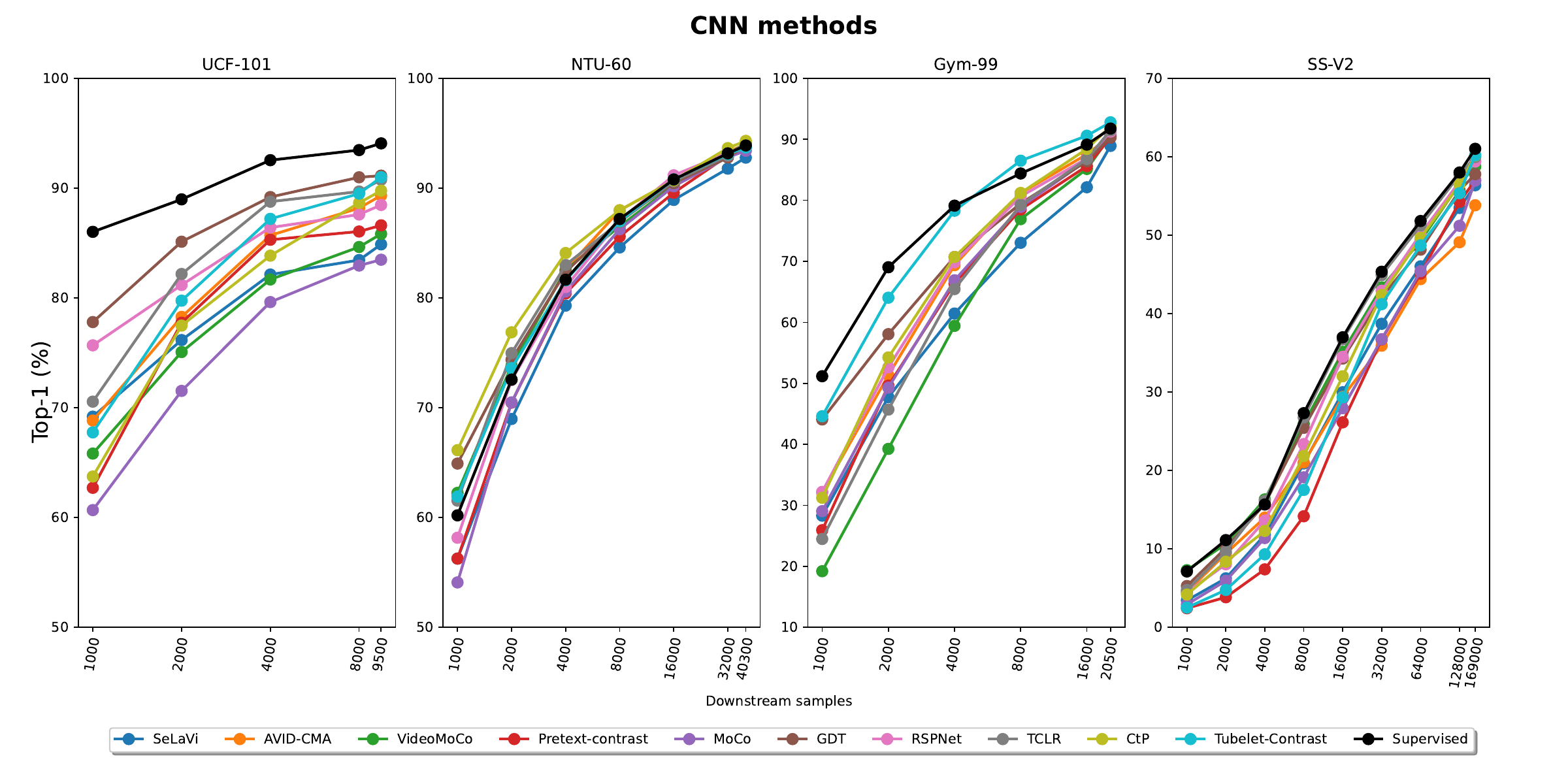}
    \includegraphics[width=0.8\linewidth]{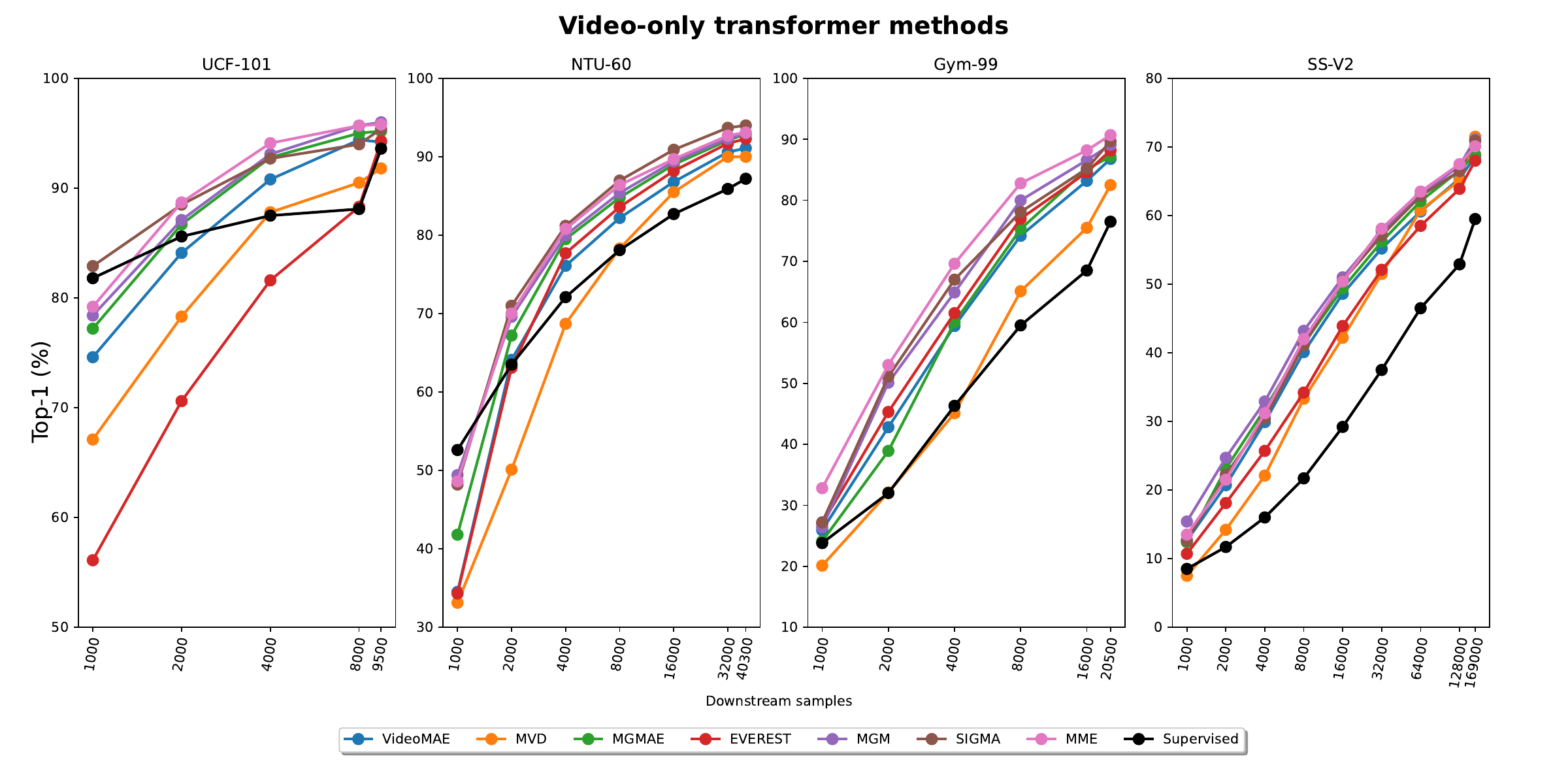}
    \includegraphics[width=0.8\linewidth]{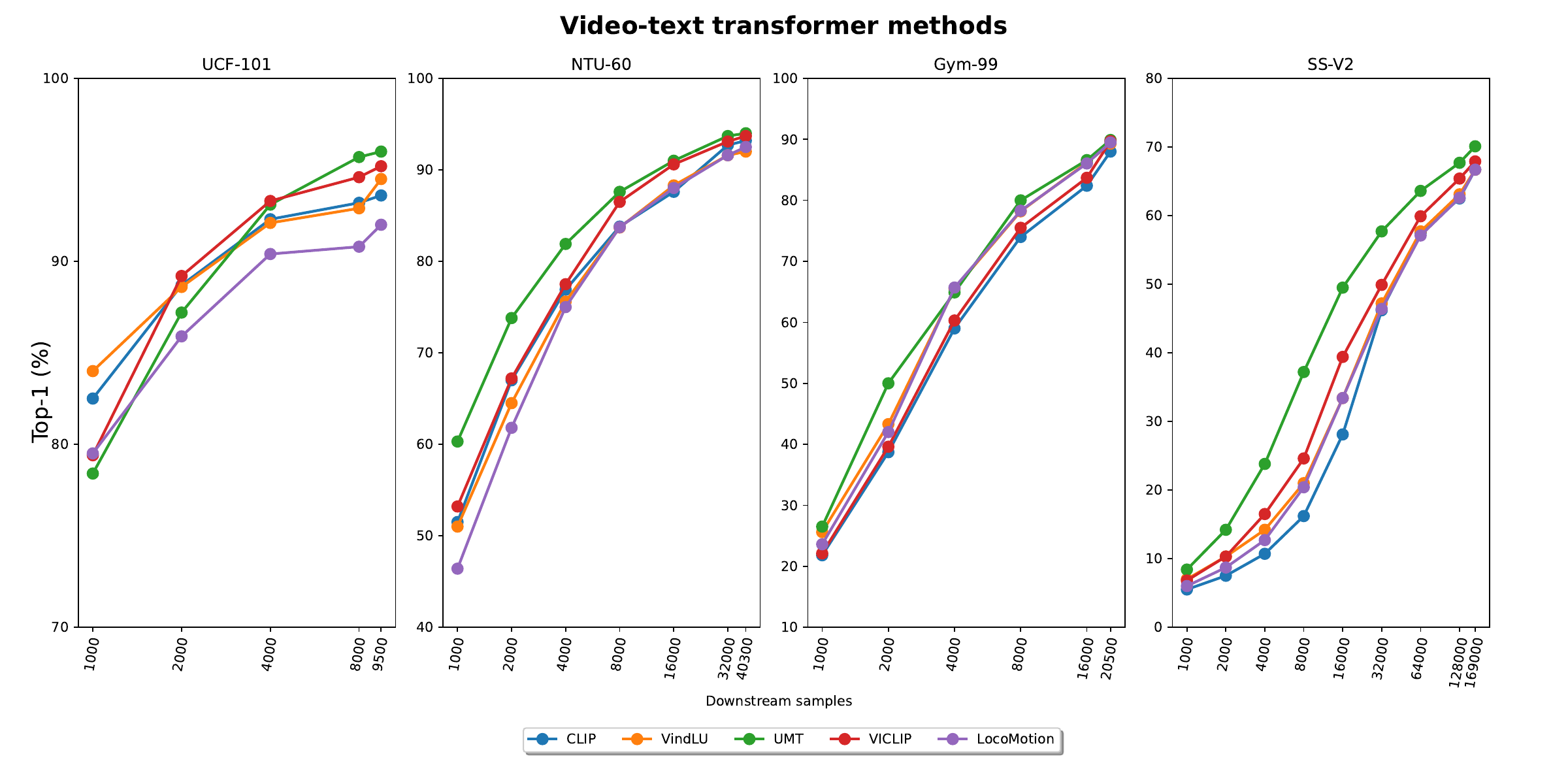}
    \caption{\textbf{Factor II: Sensitivity to Downstream Samples.} Comparison of video representation learning methods (CNNs top row, video-only Transformers middle row, and video-language Transformers bottom row)  using varying numbers of finetuning samples for four downstream datasets. Both the gap and rank among pre-training methods are sensitive to the number of samples available for finetuning. With sufficient finetuning data, the performance gap between different methods is narrow.
    }
    \label{fig:training-data-size}
\end{figure*}
\clearpage}

\medskip
\noindent\textbf{CNNs.} We first observe that the trends in the low data regime are different from those with the full data. The gap between supervised and self-supervised pre-training is much larger in low-data settings, particularly for UCF-101 and Gym-99. NTU is an exception, where, with 1000-4000 samples, CtP, GDT, AVID-CMA, and TCLR outperform supervised pre-training. 
As with changes in the downstream domain, change in the amount of downstream examples also causes a affects in the ranking of self-supervised models. For example, on UCF-101, RSPNet is much more successful than CtP and TCLR  when using only 1000 samples. 
 This is because some self-supervised models benefit more than others from an increased amount of downstream samples. For example, CtP is one of the most generalizable pre-training strategies when finetuning with the full data on UCF-101, Gym-99, and SS-v2, but this is not the case with fewer training samples. 
Interestingly, GDT is consistently high in the ranking with low amounts of finetuning samples. This is likely due to its large number of temporal augmentations, which help the generalization ability when the training data is limited.\\
\noindent\textbf{Video-only Transformers.} 
The trends observed with the full data are also different from those with the low data regimes for video-only transformer models. Both the rankings of models and the performance gaps between them changes across the different number of finetuning samples used. This is most noticable with supervised pre-training which is the worst performing model with the full data, but one of the best for UCF-101 and NTU-60 with only 1000 finetuning examples. 
This is most likely due to the low domain shift and high label overlap of UCF-101 and NTU-60 with the Kinetics-400 pre-training dataset, which benefits supervised pre-training because it is already trained for classification with Kinetics-400 labels. For GYM-99 and SS-v2 with low label overlap and a high domain shift, the performance of supervised pre-training is much lower for low-shot settings. 

The performance among video self-supervised methods also vary significantly for different amounts of downstream samples. The performance gap between the methods is much wider in the low shot setting than in the full dataset setting.
MVD is consistently lower across all settings except for UCF-101, where EVEREST has the lowest performance. There is no model which is the best across the low-data settings for different datasets,  
 again suggesting that high performance with a large amount of finetuning data does not guarantee generalization to low-shot settings. While there is no best method, MME, SIGMA and MGM consistently rank highly. These methods were also the most robust to downstream domain and their strong performance on low-data settings further suggest that non-pixel targets (MME, SIGMA) and motion-focused masking (MME, MGM) result in more generalizable representations.
Compared to CNN-based approaches, video-only transformer approaches are less generalizable to low-data settings. For instance, with 1000 finetuning samples, the best performing CNN method achieves 66.1 on NTU-60 and 44.6 on Gym-99, while the best performing video-only transformer method achieves 49.4 and 32.8.

\noindent\textbf{Video-text Transformers.} 
Compared to CNNs and video-only transformers we observe fewer rank changes in video-text transformer models across different finetuning settings, with UMT being the best method in almost all scenarios. There are exceptions to this, \eg~ in UCF-101 UMT drops from the best method with the full-finetuning data to the worst with 1000 examples. However, for the other three datasets the best performing methods with the full data are also the best with other data settings. The gap between video-text transformer models is also more stable across different data settings. One explanation for these observations could be the much larger amount of pre-training data used by video-text tranformer models.

Next, we observe that the performance of video-text models in low-data regimes is on par or worse than both CNNs-based and video-only transformer methods. 
This is especially noticeable for low sample settings (1000-4000) from datasets with a higher domain shift \ie, SSv2 and GYM99. This again validates a lack of strong generalization capability in video representations learned by video-text transformer methods.
Third, unlike with CNNs and video-only transformer methods, the ranking of video-language models is relatively stable across various amounts of downstream samples, with UMT being the best model in almost all scenarios.

\begin{myboxi}[]{red!30}
\textbf{Conclusion.}
We observe from  \cref{fig:training-data-size} that video-only  self-supervised models (both CNN and transformers) are highly sensitive to the amount of samples available for finetuning, with both the gap and rank between methods changing considerably across sample sizes on each dataset. Furthermore, CNN-based methods often outperform transformer-based approaches in low-data settings. 
Overall, there is little correlation between the performances in high- and low-data regimes, highlighting the need for standard benchmarking of video representation learning methods on this downstream factor.
\end{myboxi}
\section{Sensitivity Factor III: Downstream Actions}
\label{sec:factor_3}
As indicated earlier, existing evaluations of video representation learning methods have been limited to coarse-grained action recognition.
In this section, we investigate whether current video representation learning methods are only effective for these types of benchmarks or whether they are able to learn features that are useful for differentiating more challenging and semantically similar actions.

FineGym~\cite{Gym-99-arxiv} provides us with an experimental setup to study sensitivity to this factor. The dataset contains different evaluations with varying levels of semantic similarity, namely action recognition \textit{across all events}, \textit{within an event} or \textit{within a set}.
Recognition \textit{across all events} uses the whole of Gym-99 containing actions from four gymnastic events. For recognition \textit{within an event} there are two subsets: Vault and Floor containing only actions from these two events. Recognition \textit{within a set} has two subsets namely FX-S1, containing different \textit{leaps-jumps-hops} in Floor, and UB-S1, which consists of types of \textit{circles} in Uneven Bars. We also experiment with the long-tailed version of FineGym, Gym-288, which adds 189 more tail classes. Details of these subsets are in the appendix. As before, we attach a classification head to the pre-trained models and finetune the whole network with the training set of each subset. We report Top-1 accuracy (mean per-class) on the testing sets following \cite{Gym-99-arxiv}. 

\begin{table}[t]
\centering
\midsepremove
\captionsetup{font=small,skip=2mm}
\caption[]{\textbf{Factor III: Sensitivity to Downstream Actions Granularities for CNN methods.} Video self-supervised models evaluated on different semantic similarities of action in FineGym: across events, within an event and within a set. Colors denote relative rankings across methods for each dataset, ranging from \textcolor{lowcolor}{low}
\begin{tikzpicture}%
    \pgfplotscolorbardrawstandalone[%
        colormap name=PiYG,%
        colorbar horizontal,%
        colorbar style={%
            height=0.18cm,%
            width=2cm,%
            hide axis,%
        }%
    ]%
\end{tikzpicture} 
\textcolor{highcolor}{high}. Many methods struggle on the within a set benchmark where actions are most semantically similar.}

\setlength{\tabcolsep}{0.4mm}
{\fontsize{4.5pt}{10pt}\selectfont
\begin{tabular}{l\C{84.8}{88.9}c\C{25.4}{36.9}\C{76.0}{87.7}c\C{51.0}{80.1}\C{80.9}{91.0}c\C{52.8}{57.4}}
\toprule
\addlinespace[0.04cm]
 & \multicolumn{7}{c}{\textbf{Gym99}} & & \multicolumn{1}{c}{\textbf{Gym288}} \\ 
\addlinespace[0.04cm]
\cmidrule{2-8}\cmidrule{10-10} 
\addlinespace[0.04cm]
\multicolumn{1}{l}{\textbf{Pre-training}} & \multicolumn{1}{c}{Across Events} 
& & \multicolumn{2}{c}{Within Event} & & \multicolumn{2}{c}{Within Set} & & \multicolumn{1}{c}{Across Events}\\ 
\addlinespace[0.04cm]
\cmidrule{2-2}\cmidrule{4-5}\cmidrule{7-8}\cmidrule{10-10}
\addlinespace[0.04cm]
\multicolumn{1}{c}{} & \multicolumn{1}{c}{All} && \multicolumn{1}{c}{Vault} & \multicolumn{1}{c}{Floor} && \multicolumn{1}{c}{FX-S1} & \multicolumn{1}{c}{UB-S1} &  &  \multicolumn{1}{c}{All} \\ 
\addlinespace[0.02cm]
\arrayrulecolor{black}\midrule
\addlinespace[0.01cm]
         None                    & \multicolumn{1}{c}{84.8} && \multicolumn{1}{c}{24.7}  & \multicolumn{1}{c}{75.9} && \multicolumn{1}{c}{46.6} & \multicolumn{1}{c}{82.3} && \multicolumn{1}{c}{50.0}  \\
        \addlinespace[0.01cm]\midrule        \addlinespace[0.01cm]
SeLaVi              & 84.5 && 25.4 & 76.0 && 51.3 & 80.9 && 52.8 \\
AVID-CMA            & 85.7 && 30.4 & 82.7 && 68.0 & 87.3 && 52.5 \\
VideoMoCo           & 85.9 && 28.4 & 79.5 && 57.3 & 83.9 && 54.1 \\
Pretext-contrast    & 86.0 && 28.5 & 81.4 && 66.1 & 86.1 && 52.7 \\
MoCo                & 86.5 && 33.2 & 83.3 && 65.0 & 84.5 && 55.1 \\
GDT                 & 86.6 && 36.9 & 83.6 && 66.0 & 83.4 && 55.4 \\
RSPNet              & 86.9 && 33.4 & 82.7 && 65.4 & 83.6 && 55.2 \\
TCLR                & 87.7 && 29.8 & 84.3 && 60.7 & 84.7 && 55.4 \\
CtP                 & 88.1 && 26.8 & 86.2 && 79.1 & 88.8 && 56.5 \\
Tubelet-Contrast    & 88.9 && 28.4 & 87.7 && 80.1 & 91.0 && 57.4 \\
        \addlinespace[0.01cm]\midrule        \addlinespace[0.01cm]
Supervised          & \multicolumn{1}{c}{88.6} && \multicolumn{1}{c}{37.7} & \multicolumn{1}{c}{86.1} && \multicolumn{1}{c}{79.0} & \multicolumn{1}{c}{87.1} & & \multicolumn{1}{c}{58.4} \\
        \addlinespace[0.01cm]
\bottomrule
\end{tabular}%
}

\label{granularity_cnn}
\end{table}

\begin{table}[t!]
\centering
    \midsepremove
\captionsetup{font=small,skip=2mm}
\caption[]{\textbf{Factor III: Sensitivity to Downstream Actions Granularities for video-only Transformer methods.} Video self-supervised models evaluated on different semantic similarities of action in FineGym: across events, within an event and within a set. 
The performances of SSL methods are not strongly correlated across semantically similar action granularities.}
\setlength{\tabcolsep}{0.4mm}
{\fontsize{5pt}{10pt}\selectfont
\begin{tabular}{l\C{73.8}{85.7}@{\hskip 2mm}c\C{21.6}{25.1}\C{58.6}{80.4}@{\hskip 2mm}c\C{31.3}{57.0}\C{50.5}{88.0}c\C{36.5}{48.6}}
\toprule
\addlinespace[0.04cm]
 & \multicolumn{7}{c}{\textbf{Gym99}} & & \multicolumn{1}{c}{\textbf{Gym288}} \\ 
\addlinespace[0.04cm]
\cmidrule{2-8}\cmidrule{10-10} 
\addlinespace[0.04cm]
\multicolumn{1}{l}{\textbf{Pre-training}} & \multicolumn{1}{c}{Across Events} 
& & \multicolumn{2}{c}{Within Event} && \multicolumn{2}{c}{Within Set} & &  \multicolumn{1}{c}{Across Events}\\ 
\addlinespace[0.04cm]
\cmidrule{2-2}\cmidrule{4-5}\cmidrule{7-8}\cmidrule{10-10}
\addlinespace[0.04cm]
\multicolumn{1}{c}{} & \multicolumn{1}{c}{All} && \multicolumn{1}{c}{Vault} & \multicolumn{1}{c}{Floor} && \multicolumn{1}{c}{FX-S1} & \multicolumn{1}{c}{UB-S1} &  &  \multicolumn{1}{c}{All} \\ 
\addlinespace[0.02cm]
\arrayrulecolor{black}\midrule
\addlinespace[0.01cm]
         None                    & \multicolumn{1}{c}{34.5} && \multicolumn{1}{c}{16.7}  & \multicolumn{1}{c}{14.9} && \multicolumn{1}{c}{10.5} & \multicolumn{1}{c}{11.3} && \multicolumn{1}{c}{14.0}  \\
        \addlinespace[0.01cm]\midrule        \addlinespace[0.01cm]
VideoMAE       & 73.8 && 21.6 & 71.3 && 42.8 & 65.3 &&  41.6  \\
MVD            & 74.6 && 25.1 & 58.6 &&  31.3& 50.5 &&  36.5  \\
MGMAE          & 80.9 && 23.9 & 69.8 && 33.7 & 79.5 &&  41.7  \\
EVEREST        & 81.9 && 24.9 & 71.7 && 39.0 & 88.0 && 44.7 \\
MGM & 83.7 && 21.6 & 76.2 && 38.6 & 86.9 && 46.8 \\
SIGMA & 84.4 && 23.1 & 77.7 && 55.1 & 79.9 && 47.4 \\
MME            & 85.7 && 21.7 & 80.4 && 57.0 & 91.2 &&  48.6  \\
        \addlinespace[0.01cm]\midrule        \addlinespace[0.01cm]
Supervised & \multicolumn{1}{c}{68.1} && \multicolumn{1}{c}{26.4} & \multicolumn{1}{c}{54.6} && \multicolumn{1}{c}{35.7} & \multicolumn{1}{c}{63.1} &&  \multicolumn{1}{c}{32.9}  \\
\bottomrule
\end{tabular}%
}
\label{granularity_video}
\end{table}

\begin{table}[t!]
\centering
    \midsepremove
\captionsetup{font=small,skip=2mm}
\caption[]{\textbf{Factor III: Sensitivity to Downstream Actions for video-text Transformer methods.} 
Similar to video-only SSL models, video-text models are also  less  correlated across different action granularities.}
\setlength{\tabcolsep}{0.4mm}
{\fontsize{5pt}{10pt}\selectfont
\begin{tabular}{l\C{82.8}{84.9}@{\hskip 2mm}c\C{27.1}{28.3}\C{69.8}{82.9}@{\hskip 2mm}c\C{48.0}{68.0}\C{49.4}{68.0}c\C{46.3}{49.5}}
\toprule
\addlinespace[0.04cm]
 & \multicolumn{7}{c}{\textbf{Gym99}} & & \multicolumn{1}{c}{\textbf{Gym288}} \\ 
\addlinespace[0.04cm]
\cmidrule{2-8}\cmidrule{10-10} 
\addlinespace[0.04cm]
\multicolumn{1}{l}{\textbf{Pre-training}} & \multicolumn{1}{c}{Across Events} 
& & \multicolumn{2}{c}{Within Event} && \multicolumn{2}{c}{Within Set} & &  \multicolumn{1}{c}{Across Events}\\ 
\addlinespace[0.04cm]
\cmidrule{2-2}\cmidrule{4-5}\cmidrule{7-8}\cmidrule{10-10}
\addlinespace[0.04cm]
\multicolumn{1}{c}{} & \multicolumn{1}{c}{All} && \multicolumn{1}{c}{Vault} & \multicolumn{1}{c}{Floor} && \multicolumn{1}{c}{FX-S1} & \multicolumn{1}{c}{UB-S1} &  &  \multicolumn{1}{c}{All} \\ 
\addlinespace[0.02cm]
\arrayrulecolor{black}\midrule
\addlinespace[0.01cm]
None & \multicolumn{1}{c}{34.5} & & \multicolumn{1}{c}{16.7} & \multicolumn{1}{c}{14.9} && \multicolumn{1}{c}{19.5} & \multicolumn{1}{c}{11.3} && \multicolumn{1}{c}{14.0} \\
\midrule
         CLIP & 82.8 && 31.9 & 69.8 && 48.0 & 49.4 &&  46.3  \\
         VindLU & 84.2 && 28.3  &79.8  && 59.6  & 66.7 &&  47.8  \\
         Locomotion & 84.2 && 32.0 & 80.6 && 59.0 & 68.5 &&  48.0  \\
         UMT & 84.6 && 27.6 & 82.9 && 68.0 & 68.0 &&  48.0  \\
         VICLIP & 84.9 && 27.1 & 77.2 && 57.3 & 60.7 &&  49.5  \\
\bottomrule
\end{tabular}%
}
\label{granularity_vl}
\end{table}

\noindent\textbf{CNNs.} From Table~\ref{granularity_cnn}, we observe that the performance of self-supervised methods varies considerably across downstream actions. The methods that perform best on Gym-99 often do not generalize well to the subsets with higher semantic similarity among actions. This is particularly noticeable for RSPNet and TCLR, which drop in the ranking for the within-set subsets. All self-supervised methods, except GDT, struggle on Vault, likely due to the intense motions. 
Surprisingly, MoCo performs reasonably well when actions are more semantically similar, and is comparable to GDT and RSPNet.  
The best self-supervised methods for subsets with high semantic similarity are CtP and Tubelet-Contrast, even outperforming supervised pre-training on FX-S1 and UB-S1. This is especially evident from FX-S1, where they outperform the next-best self-supervised method, AVID-CMA, by more than 12\%. As with downstream domain and samples, supervised pre-training generalizes better than self-supervised methods across downstream actions, with only Tubelet-contrast and CtP achieving comparable performance. 

\cref{granularity_cnn} also compares balanced Gym-99 with long-tailed Gym-288. We observe that 
the ranking remains consistent, meaning the performance on the balanced set is generally indicative of the performance on the long-tailed set. 

\noindent\textbf{Video-only Transformers.} Table~\ref{granularity_video} shows the performance of video-only transformer-based methods for different action granularities. Similar to CNN methods, we observe a significant change in ranking among the methods for subsets with varying action granularity. This is particularly noticeable for the lower-ranked methods on Gym-99, \eg~VideoMAE is worst at Gym-99, but third on FX-S1. However, the top two methods for Gym-99 \ie, SIGMA and MME seem to generalize well across different subsets except Vault. 
Notably, for the most fine-grained granularities, FX-S1 and UB-S1, the performance difference between the best and worst methods is much higher than other granularities, thus indicating a more substantial challenge for SSL evaluation.  Overall, this demonstrates that different methods learn features suitable for different action granularities, and a strong performance on one granularity does not guarantee the same for a different action granularity. 


As with CNNs, the performance of balanced Gym-99 with long-tailed Gym-288 seems co-related, with no significant rank change. We also observe that supervised training performs the worst for most subsets.
Finally, comparing the results with Table~\ref{granularity_cnn}, we observe that the best CNN methods significantly outperform top video-only SSL methods across all granularities. This again highlights the need for a more varied standard benchmark as current work has converged on MAE-based transformers which perform poorly for fine-grained actions.

\noindent\textbf{Video-text Transformers.}
~\cref{granularity_vl} shows that similar to the CNNs and video-only transformer methods, video-text methods also perform variably with different action granularities. For example, UMT achieves the best performance for FX-S1 and Floor but observes suboptimal performance for Vault. LocoMotion performs best on Vault and UB-S1, but performs sub-optimally on FX-S1.  We observe no strong correlations between performances across different granularities for each method.

Overall, Vault has a very different ranking for all three method types. This is potentially due to very intense motions, which most methods struggle to encode. Surprisingly, CLIP and MoCo are two of the best models on Vault, meaning image-based methods that ignore motion are potentially more robust to intense motions than current video encoders.  Video-text methods seem to observe a substantial gain over many video-only transformer methods for various granularities, \eg, Vault and FX-S1, however, CNN-based approaches, including supervised pre-training, are the best overall. 

\begin{myboxi}[]{NavyBlue!30}
\textbf{Conclusion.}
All three categories of evaluated video representation methods are sensitive to the actions present in the downstream dataset and do not generalize well across diverse granularities of actions. 
The performance gap between different methods is much greater than that for domain-shift evaluation. This emphasizes the need to benchmark self-supervised and video-language methods beyond coarse-grained actions and domains to more subtle action granularities.

\end{myboxi}
\section{Sensitivity Factor IV: Downstream Tasks}
\label{sec:factor_4}
The fourth factor we investigate is whether video representation learning models are sensitive to the downstream task or whether features learned by such methods are useful to video understanding tasks beyond action recognition. 
We evaluate this in two ways. First, we keep the domain fixed and evaluate different tasks within the same downstream domain.  Second, we also explore further tasks by changing the downstream domain and seeing how these two factors interplay.

\begin{table*}[t]
\centering
    \midsepremove
\captionsetup{font=small,skip=2mm}
\caption[]{\textbf{Factor IV: Sensitivity to Downstream Tasks for CNN methods.} Transferability of self-supervised video learning methods across video understanding tasks. Colors denote relative rankings across methods for each task, ranging from \textcolor{lowcolor}{low} \begin{tikzpicture}%
      \pgfplotscolorbardrawstandalone[%
        colormap name=PiYG,%
        colorbar horizontal,%
        colorbar style={%
          height=0.18cm,%
          width=2cm,%
          hide axis,%
        }%
      ]%
    \end{tikzpicture} \textcolor{highcolor}{high}. Note that for repetition counting, lower (error) is better.  
    Self-supervised features are transferable to different downstream tasks when the domain shift is low, but struggle when there is also a domain shift. Action recognition on UCF-101 is not a good proxy for self-supervised video learning use cases where a downstream domain- and task-shift can be expected. 
    }
\setlength{\tabcolsep}{3mm}
\resizebox{\linewidth}{!}{%
\begin{tabular}{l\C{83.3}{91.0}\C{0.416}{0.476}\CR{0.123}{0.208}\C{72.9}{87.0}\C{31.9}{49.5} c\C{8.2}{12.2}\C{10.0}{14.1}\C{33.2}{35.9}}
\toprule
\addlinespace[0.07cm]
 & \multicolumn{5}{c}{\textbf{Task-shift within domain}} & & \multicolumn{3}{c}{\textbf{Task-shift out of domain}} \\ 
\addlinespace[0.04cm]
\cmidrule{2-6}\cmidrule{8-10} 
\addlinespace[0.04cm]
\addlinespace[0.04cm]
\multicolumn{1}{l}{\textbf{Pre-training}} & \multicolumn{1}{c}{Action} & \multicolumn{1}{c}{Spatio-Temporal} & \multicolumn{1}{c}{Repetition $\downarrow$} &\multicolumn{1}{c}{Arrow of}  &\multicolumn{1}{c}{Temporal } & & \multicolumn{1}{c}{Multi-label}  &  \multicolumn{1}{c}{Spatio-Temporal} & \multicolumn{1}{c}{Temporal } \\ 
\multicolumn{1}{c}{} & \multicolumn{1}{c}{Recognition} & \multicolumn{1}{c}{Action Detection} & \multicolumn{1}{c}{Counting} &\multicolumn{1}{c}{Time} &\multicolumn{1}{c}{{Action Localization}} & & \multicolumn{1}{c}{Recognition}  &  \multicolumn{1}{c}{Action Detection} & \multicolumn{1}{c}{{Action Localization}} \\ 
\addlinespace[0.02cm]
\midrule
\addlinespace[0.01cm]
None                    & \multicolumn{1}{c}{77.3} & \multicolumn{1}{c}{0.330}  & \multicolumn{1}{c}{0.217} & \multicolumn{1}{c}{56.1} & \multicolumn{1}{c}{14.1} && \multicolumn{1}{c}{7.9} & \multicolumn{1}{c}{7.4} & \multicolumn{1}{c}{30.9}  \\ 
\addlinespace[0.01cm]\midrule        \addlinespace[0.01cm]
MoCo                   & 83.3 & 0.416 & 0.208 & 80.3 &41.2 && 8.3  & 11.7 & 34.5 \\ 
VideoMoCo              & 84.9 & 0.440 & 0.185 & 72.9 &44.1 && 10.5 & 13.1 & 34.7 \\ 
SeLaVi                 & 85.2 & 0.419 & 0.162 & 77.4 &31.9 && 8.4  & 10.2 & 34.9 \\ 
Pretext-contrast       & 87.7 & 0.462 & 0.164 & 77.2 &41.0 && 8.9  & 12.7 & 34.7 \\ 
RSPNet                 & 88.7 & 0.467 & 0.145 & 87.0 &49.5 && 9.0  & 14.1 & 35.9 \\ 
AVID-CMA               & 88.8 & 0.435 & 0.148 & 83.3 &43.8 && 8.2  & 10.0 & 35.5 \\ 
CtP                    & 90.1 & 0.465 & 0.178 & 77.1 &33.5 && 9.6  & 10.0 & 33.2 \\ 
TCLR                   & 90.8 & 0.476 & 0.142 & 85.6 &32.4 && 12.2 & 10.8 & 34.3 \\ 
GDT                    & 91.3 & 0.463 & 0.123 & 76.4 &48.7 && 8.5  & 12.6 & 35.9 \\ 
Tubelet-Contrast                    & 91.0 & 0.463 & 0.150 & 79.1 &  35.5 && 9.9 & 10.3 & 35.4 \\ 
\addlinespace[0.01cm]\midrule         \addlinespace[0.01cm]
Supervised             & \multicolumn{1}{c}{93.9} & \multicolumn{1}{c}{0.482} & \multicolumn{1}{c}{0.132} & \multicolumn{1}{c}{77.0} & \multicolumn{1}{c}{60.7} && \multicolumn{1}{c}{23.5} & \multicolumn{1}{c}{17.9}  & \multicolumn{1}{c}{36.3}\\ 
\addlinespace[0.01cm]
\bottomrule
\end{tabular}%
}
\label{task_shift_cnn}
\end{table*}

\begin{table*}[t]
\centering
    \midsepremove
\captionsetup{font=small,skip=2mm}
\caption[]{\textbf{Factor IV: Sensitivity to Downstream Tasks for video-only Transformer methods.} Transferability of self-supervised video learning methods across video understanding tasks. Colors denote relative rankings across methods for each task, ranging from \textcolor{lowcolor}{low} \begin{tikzpicture}%
      \pgfplotscolorbardrawstandalone[%
        colormap name=PiYG,%
        colorbar horizontal,%
        colorbar style={%
          height=0.18cm,%
          width=2cm,%
          hide axis,%
        }%
      ]%
    \end{tikzpicture} \textcolor{highcolor}{high}. 
Note that for repetition counting, lower (error) is better.
Video-only SSL methods show a good transferability to various video understanding tasks within the same domain and out of domain, with top methods demonstrating a good correlation.
}
\setlength{\tabcolsep}{3mm}
\resizebox{\linewidth}{!}{%
\begin{tabular}{l\C{93.3}{96.0}\C{0.762}{0.793}\CR{0.152}{0.184}\C{90.1}{98.9}\C{56.3}{62.7}c\C{14.4}{23.6}\C{22.0}{27.3}\C{36.9}{37.8}}
\toprule
\addlinespace[0.07cm]
 & \multicolumn{5}{c}{\textbf{Task-shift within domain}} & & \multicolumn{3}{c}{\textbf{Task-shift out of domain}} \\ 
\addlinespace[0.04cm]
\cmidrule{2-6}\cmidrule{8-10} 
\addlinespace[0.04cm]
\addlinespace[0.04cm]
\multicolumn{1}{l}{\textbf{Pre-training}} & \multicolumn{1}{c}{Action} & \multicolumn{1}{c}{Spatio-Temporal} & \multicolumn{1}{c}{Repetition $\downarrow$} &\multicolumn{1}{c}{Arrow of}  &\multicolumn{1}{c}{Temporal } & & \multicolumn{1}{c}{Multi-label}  &  \multicolumn{1}{c}{Spatio-Temporal} & \multicolumn{1}{c}{Temporal } \\ 
\multicolumn{1}{c}{} & \multicolumn{1}{c}{Recognition} & \multicolumn{1}{c}{Action Detection} & \multicolumn{1}{c}{Counting} &\multicolumn{1}{c}{Time} &\multicolumn{1}{c}{{Action Localization}} & & \multicolumn{1}{c}{Recognition}  &  \multicolumn{1}{c}{Action Detection} & \multicolumn{1}{c}{{Action Localization}} \\ 
\addlinespace[0.02cm]
\midrule
\addlinespace[0.01cm]
None                    & \multicolumn{1}{c}{52.1} & \multicolumn{1}{c}{0.570}  & \multicolumn{1}{c}{0.451} & \multicolumn{1}{c}{00.0} & \multicolumn{1}{c}{11.5} && \multicolumn{1}{c}{8.6} & \multicolumn{1}{c}{9.5} & \multicolumn{1}{c}{30.3}  \\
\addlinespace[0.01cm]\midrule        \addlinespace[0.01cm]
EVEREST           & 93.3 & 0.789 & 0.174 & 93.6 &  57.3 && 17.8  &  24.8 &  36.9 \\
MVD               & 94.0 & 0.762 & 0.184 & 90.1 &  59.3 && 16.1  & 22.0 &    37.6 \\
VideoMAE          & 94.2 & 0.788 & 0.172 & 97.8 &  58.6 && 14.4 & 26.6 &       37.3 \\
MGMAE    & 95.2 & 0.793   & 0.181 & 96.8 &         56.3 && 17.9  & 26.9 &    37.3 \\
SIGMA & 95.4 & 0.793 & 0.178 & 94.0 &              62.7 && 22.4  & 27.3 &    37.8 \\
MME               & 95.8 & 0.793  & 0.155 & 98.9 & 61.8 && 23.6  & 26.6 &    37.4 \\
MGM & 96.0 & 0.788 & 0.152 & 98.2 &                62.0 && 22.5  & 27.3 &     37.6 \\
\addlinespace[0.01cm]\midrule         \addlinespace[0.01cm]
Supervised & \multicolumn{1}{c}{93.6} & \multicolumn{1}{c}{0.761} & \multicolumn{1}{c}{0.381} & \multicolumn{1}{c}{98.2} & \multicolumn{1}{c}{60.9} && \multicolumn{1}{c}{17.3} & \multicolumn{1}{c}{18.1} & \multicolumn{1}{c}{36.9}\\
\bottomrule
\end{tabular}%
}
\label{task_shift_video_only}
\end{table*}

\begin{table*}[t]
\centering
    \midsepremove
\captionsetup{font=small,skip=2mm}
\caption[]{\textbf{Factor IV: Sensitivity to Downstream Tasks for video-text Transformer methods.} Transferability of self-supervised video learning methods across video understanding tasks. Colors denote relative rankings across methods for each task, ranging from \textcolor{lowcolor}{low} \begin{tikzpicture}%
      \pgfplotscolorbardrawstandalone[%
        colormap name=PiYG,%
        colorbar horizontal,%
        colorbar style={%
          height=0.18cm,%
          width=2cm,%
          hide axis,%
        }%
      ]%
    \end{tikzpicture} \textcolor{highcolor}{high}. 
    Note that for repetition counting, lower (error) is better.
    Video-text models have more or less similar rankings across different tasks, with a large gap between best and worst methods.
     }
\setlength{\tabcolsep}{3mm}
\resizebox{\linewidth}{!}{%
\begin{tabular}{l\C{93.6}{96.0}\C{0.638}{0.723}\CR{0.320}{0.521}\C{50.1}{89.5}\C{48.8}{65.5}c\C{34.9}{44.9}\C{8.8}{22.3}\C{36.0}{38.0}}
\toprule
\addlinespace[0.07cm]
 & \multicolumn{5}{c}{\textbf{Task-shift within domain}} & & \multicolumn{3}{c}{\textbf{Task-shift out of domain}} \\ 
\addlinespace[0.04cm]
\cmidrule{2-6}\cmidrule{8-10} 
\addlinespace[0.04cm]
\addlinespace[0.04cm]
\multicolumn{1}{l}{\textbf{Pre-training}} & \multicolumn{1}{c}{Action} & \multicolumn{1}{c}{Spatio-Temporal} & \multicolumn{1}{c}{Repetition $\downarrow$} &\multicolumn{1}{c}{Arrow of}  &\multicolumn{1}{c}{Temporal } & & \multicolumn{1}{c}{Multi-label}  &  \multicolumn{1}{c}{Spatio-Temporal} & \multicolumn{1}{c}{Temporal } \\ 
\multicolumn{1}{c}{} & \multicolumn{1}{c}{Recognition} & \multicolumn{1}{c}{Action Detection} & \multicolumn{1}{c}{Counting} &\multicolumn{1}{c}{Time} &\multicolumn{1}{c}{{Action Localization}} & & \multicolumn{1}{c}{Recognition}  &  \multicolumn{1}{c}{Action Detection} & \multicolumn{1}{c}{{Action Localization}} \\ 
\addlinespace[0.02cm]
\midrule
\addlinespace[0.01cm]
None                    & \multicolumn{1}{c}{52.1} & \multicolumn{1}{c}{0.570}  & \multicolumn{1}{c}{0.451} & \multicolumn{1}{c}{0.0} & \multicolumn{1}{c}{11.5} && \multicolumn{1}{c}{8.6} & \multicolumn{1}{c}{9.5} & \multicolumn{1}{c}{30.3}  \\
\addlinespace[0.01cm]\midrule        \addlinespace[0.01cm]
CLIP         & 93.6 & 0.638 & 0.520 & 50.1 & 48.8 && 34.9 & 8.9 & 36.0\\
LocoMotion       & 92.0 & 0.645 & 0.490 & 53.1 & 59.5   && 35.0 & 9.3 & 37.5 \\
VindLU      & 94.5 & 0.641 & 0.490 & 53.0 &60.1 && 37.7 & 9.4 & 37.9\\
VICLIP       & 95.2 & 0.674 & 0.450 & 89.5 &  59.6&& 38.9 & 14.9 & 36.7\\
UMT    & 96.0 & 0.723 & 0.321 & 57.5 & 65.5 && 44.9 & 22.3 & 37.0\\
\bottomrule
\end{tabular}%
}
\label{task_shift_video_language}
\end{table*}

\subsubsection{Task-shift within domain.}
We consider four different tasks which are all defined for the UCF domain: spatio-temporal action detection~\cite{yowo}, repetition counting~\cite{rep_counting}, arrow-of-time prediction~\cite{arrow_of_time}, and temporal action localization~\cite{THUMOS14}. Using UCF allows us to keep the domain fixed across tasks and eliminates the impact of domain shift. Note that each task uses a different setup from UCF-101 action recognition, however, the domain remains consistent. 
\subsubsection{Task-shift out of domain.} 
We also aim to evaluate the models on setups with both the domain and the task change. We do so with three popular video understanding benchmarks: long-term multi-label classification on Charades \cite{charades-sigurdsson:hal-01418216}, short-term spatio-temporal action detection on AVA \cite{AVA-Gu_2018_CVPR} and temporal action localization on ActivityNet~\cite{caba2015activitynet}. Note that each dataset belongs to a different domain and is defined for different tasks. 

For each task, we use the R(2+1)D-18 or ViT-B networks as the pre-trained backbones as before and attach task-dependent heads. We report mean absolute counting error for repetition counting~\cite{rep_counting}, classification accuracy for arrow-of-time prediction~\cite{arrow_of_time, aot2-wei2018learning}, and mean Average Precision for temporal action localization~\cite{THUMOS14}, spatio-temporal localization~\cite{mettes2016spot} and multi-label classification~\cite{charades-sigurdsson:hal-01418216}. Further details are in the appendix.


\noindent\textbf{CNNs.}
The left part of \cref{task_shift_cnn} shows the results of CNN-based SSL methods on different tasks without any domain shift. We observe that self-supervised learning is beneficial to tasks beyond action recognition, with almost all methods outperforming training from scratch on spatio-temporal action detection, repetition counting, arrow-of-time prediction, and temporal action localization tasks. Action recognition correlates well with the spatio-temporal action detection results, but is less correlated with temporal action localization results. Repetition counting and arrow-of-time also show less correlation with action recognition, suggesting that the  UCF-101 action recognition benchmark does not indicate how well self-supervised methods generalize to other tasks. For example, RSPNet and GDT generalize the best across various tasks, while CtP ranks high on action recognition and spatio-temporal detection but performs modestly for repetition counting and temporal action localization.  For repetition counting and arrow-of-time prediction, some methods perform comparably to or outperform supervised pre-training. Overall, the results show that different methods have different task sensitivity even without any domain shift. 

The right part of \cref{task_shift_cnn} shows the results on different tasks from different domains.
Again, we observe no strong correlations between performance on different tasks for all SSL methods. All self-supervised methods struggle considerably when both the domain and task change. Supervised pre-training is significantly better on multi-label recognition and spatio-temporal action detection than all SSL methods.   Overall, the results show SSL pre-training is very susceptible to the task shift out of domain.

\noindent\textbf{Video-only Transformers.}
\cref{task_shift_video_only} shows the results of video-only SSL methods for tasks shift within and out of domain. 
We observe a mild correlation in performance for task shift within the domain. Action recognition performance is well correlated to spatio-temporal action detection, mildly correlated with arrow of time prediction, but not with repetition counting and temporal action localization. For the task shift out of domain, most methods perform similarly on temporal action localization and spatio-temporal action detection to their within-domain counterpart. This is due to the transformer-based models being more robust to domain shift than CNN-based methods. 

In contrast to CNN methods, where no method showed any strong generalization across all the evaluated tasks, top video-only transformer methods, MGM and MME, consistently perform better than the other methods across most tasks.   
Note that this is similar to the downstream domain (\cref{sec:factor_1}) where we also observe a better correlation across the different domains for top video-only transformer methods. Also, comparing tables ~\ref{task_shift_cnn} and ~\ref{task_shift_video_only} shows that the transformer methods are consistently better than top CNN methods across all task shifts except for repetition counting, where CNNs obtain a better performance. 

Similar to the prior three factors, supervised pre-training is among the worst performers and is consistently outperformed by most SSL methods. Overall, this suggests that supervised pre-training with Kinetics 400 is not an optimal learning strategy for video-only transformers.

\noindent\textbf{Video-text Transformers.}
\cref{task_shift_video_language} shows the results of video-text transformer methods for tasks shift within and out-of-domain. Video-text models have similar rankings across different tasks. Among the methods, UMT seems to be the best across most tasks, with the exception of Arrow of Time.  The gap between methods can also be very large, for instance in Arrow of Time VICLIP obtains 89.5\% compared to 50-60\% from the other models.

We observe that video-text models are generally worse than top video-only methods, except for multi-label recognition. 
This is especially visible in spatio-temporal action detection, repetition, and arrow of time prediction.  
However, for multi-label recognition, the performance gap is huge, likely due to having multiple object captions in the pre-training.  


\begin{myboxi}[]{Goldenrod!60}



For the three model types, action classification performance correlates differently with task shifts. For CNN-based video SSL methods, action recognition is not indicative of task-shift performance, especially in combination with domain shifts. Video-only transformers exhibit mild correlations within and out of domain, with 1-2 methods generalizing across different tasks.  
For video-text models action recognition models do correlate with task shifts but these models generally underperform compared to video-only models. 

\end{myboxi}
\section{SEVERE-benchmark++}

\begin{table*}[t]
    \centering
\captionsetup{font=small,skip=2mm}
\caption[]{\textbf{SEVERE-benchmark++: Performance of all three types of video representation learning methods on the proposed SEVERE-benchmark} for evaluating generalization along downstream domains, samples, actions, and tasks. VO-transformer refers to methods that only use video data for pre-training and VT-transformer refers to methods that use video-text pairs for pre-training.
}
    \resizebox{\linewidth}{!}{
    \begin{tabular}
    {
    l c \C{64.3}{82.4}c
    \C{52.0}{70.9}\C{76.5}{92.8}@{\hskip 2mm}c
    \C{56.1}{84.0}\C{19.1}{51.2}@{\hskip 2mm}c
    \C{31.3}{80.1}\C{49.4}{91.2}@{\hskip 2mm}c
    \CR{0.123}{0.520}\C{8.2}{44.9}
    }
    \toprule
    \addlinespace[0.1cm]
    & & \multicolumn{1}{c}{\textbf{Existing}} & & \multicolumn{11}{c}{\textbf{SEVERE-benchmark}} \\
    \addlinespace[0.04cm]
    \cmidrule{3-3} \cmidrule{5-15}
    \addlinespace[0.1cm]
    \multicolumn{1}{l}{\textbf{Pre-training}} & \multicolumn{1}{l}{\textbf{Method Type}} & \multicolumn{1}{c}{} & & \multicolumn{2}{c}{Domains} &  & \multicolumn{2}{c}{Samples} & & \multicolumn{2}{c}{Actions}& & \multicolumn{2}{c}{Tasks}\\
    \cmidrule(lr){5-6} \cmidrule(lr){8-9} \cmidrule(lr){11-12} \cmidrule(lr){14-15}
    \addlinespace[0.1cm]
    & & \multicolumn{1}{c}{K400} & & \multicolumn{1}{c}{SS-v2} & \multicolumn{1}{c}{Gym-99} & & \multicolumn{1}{c}{UCF ($10^{3}$)} & \multicolumn{1}{c}{Gym-99 ($10^{3}$)} & & \multicolumn{1}{c}{FX-S1 } & \multicolumn{1}{c}{UB-S1}& & \multicolumn{1}{c}{UCF-RC} & \multicolumn{1}{c}{Charades-MLC}\\
    \midrule
    \addlinespace[0.01cm]
    
        MoCo & CNN             & 64.3  && 57.1   & 90.7   && 60.6   & 29.0   && 65.0   & 84.5   && 0.208   & 8.3  \\
        VideoMoCo & CNN        & 65.0  && 59.0   & 90.3   && 65.8   & 19.1   && 57.3   & 83.9   && 0.185   & 10.5 \\
        SeLaVi & CNN           & 65.5  && 56.2   & 88.9   && 69.1   & 28.3   && 51.3   & 80.9   && 0.162   & 8.4  \\
        Pretext-Contrast & CNN & 66.1  && 56.9   & 90.5   && 62.7   & 25.9   && 66.1   & 86.1   && 0.164   & 8.9  \\
        RSPNet & CNN           & 66.4  && 59.0   & 91.1   && 75.6   & 32.2   && 65.4   & 83.6   && 0.145   & 9.0  \\
        AVID-CMA & CNN         & 66.6  && 52.0   & 90.4   && 68.8   & 32.1   && 68.0   & 87.3   && 0.148   & 8.2  \\
        CtP & CNN              & 67.1  && 59.6   & 92.0   && 63.7   & 31.2   && 79.1   & 88.8   && 0.178   & 9.6  \\
        TCLR & CNN             & 68.1  && 59.8   & 91.6   && 70.5   & 24.4   && 60.7   & 84.7   && 0.142   & 12.2 \\
        GDT & CNN              & 67.1  && 58.0   & 90.5   && 77.8   & 44.0   && 66.0   & 83.4   && 0.123   & 8.5  \\
        Tubelet-contrast & CNN & 65.8  && 60.2   & 92.8   && 67.7   & 44.6   && 80.1   & 91.0   && 0.150   & 9.9   \\
        Supervised & CNN       & -     && 60.8   & 92.1   && 86.0   & 51.2   && 79.0   & 87.1   && 0.132   & 23.5 \\
        \midrule
        EVEREST & VO-transformer   & 79.0  && 68.0   & 88.2   && 56.1   & 27.0   && 39.0   & 88.0   && 0.174   & 17.8   \\
        MVD & VO-transformer  & 79.7  && 68.5   & 82.5   && 67.1   & 20.1   && 31.3   & 50.5   && 0.184   & 16.1   \\
        MGMAE & VO-transformer  & 79.9  && 68.9   & 87.2   && 77.2   & 24.1   && 33.7   & 79.5   && 0.181   & 17.9   \\
        VideoMAE & VO-transformer    & 80.0  && 68.6   & 86.8   && 74.6   & 25.9   && 42.8   & 65.3   && 0.172   & 14.4   \\
        MME & VO-transformer    & 80.7  && 70.1   & 90.7   && 79.2   & 32.8   && 57.0   & 91.2   && 0.155   & 23.6   \\
        SIGMA & VO-transformer     & 81.3  && 70.9   & 89.7   && 82.9   & 27.2   && 55.1   & 79.9   && 0.178   & 22.4   \\
        Supervised & VO-transformer & - && 59.5 & 76.5 && 81.8 & 23.8 && 35.7 & 63.1 && 0.381 & 17.3 \\
        \midrule
        CLIP & VT-transformer & 81.8  && 66.7   & 88.0   && 82.5   & 21.8   && 48.0   & 49.4   && 0.520   & 34.9   \\
        LocoMotion & VT-transformer & 78.2  && 66.7   & 89.5   && 79.5   & 23.6   && 59.0& 68.5   && 0.490   & 35.0   \\
        VindLU & VT-transformer  & 79.1  && 66.7   & 89.3   && 84.0   & 25.6   && 59.6   & 66.7   && 0.490   & 37.7   \\
        UMT & VT-transformer    & 81.7  && 70.1   & 89.9   && 78.4   & 26.5   && 68.0   & 68.0   && 0.321   & 44.9   \\
        VICLIP & VT-transformer      & 82.4  && 67.9   & 89.7   && 79.4   & 22.1   && 57.3   & 60.7   && 0.450   & 38.9   \\
        
    \addlinespace[0.01cm]
    \bottomrule
    \end{tabular}
}
\label{proposed-benchmark}
\end{table*}

 As evident from the results in previous sections, current video self-supervised methods are benchmark-sensitive to the four factors we have studied. 
Based on our findings from the earlier version of this work, we had propose the SEVERE-benchmark (\underline{SE}nsitivity of \underline{V}id\underline{E}o \underline{RE}presentations) for use in future works to more thoroughly evaluate new video self-supervised methods for generalization along the four sensitivity factors we have examined. Since we do not expect future works to run all the experiments from our study, we create a subset of experiments that are indicative benchmarks for each sensitivity factor and realistic to run.  We summarize the benchmark composition in \cref{proposed-benchmark} and detail its motivation per factor. 
\\
\noindent\textbf{Downstream domain.} 
To measure a self-supervised model's domain sensitivity, we recommend
using Something-Something-v2 and FineGym-99. These two datasets come from domains distinct to Kinetics-400 and UCF-101 and also each other. FineGym-99 evaluates a model's ability to generalize to datasets with less distinctive backgrounds where there are few actions in common with Kinetics-400. SS-v2 evaluates the generalizability to actions that require high temporal awareness, as well as the shift to a first-person viewpoint. It is evident from \cref{proposed-benchmark} that there are significant rank changes between  Kinetics-400, Gym-99, and SS-v2 for CNNs. For video-only methods, SS-v2 shows a good correlation with Kinetics-400 performance, but Gym-99 shows a weak correlation with Kinetics-400. For video-text methods, both are mildly correlated with Kinetics-400. 
Moreover,  CNNs lag behind transformers for Kinetics-400 and SS-v2 but are better than transformers for Gym-99.  Thus, these datasets provide a challenging subset to evaluate the domain shift for future methods, both CNN-based and transformer-based. \\

\noindent\textbf{Downstream samples.} 
For the sample sensitivity, we recommend using 1000 samples on UCF-101 and Gym-99. Both show the most dramatic decoupling with Kinetics-400 performance for CNN methods. Video-only transformers show a weak correlation with Kinetics-400 for both subsets, while video-text methods are only mildly correlated. 
Furthermore, different method types have different advantages with transfromer-based approaches generally performing better on the 1000 samples from UCF-101 and best performing methods on Gym-99 (10$^3$) being CNN-based.  Therefore, these subsets can act as a challenging benchmark of sample efficiency for all three types of methods.

\noindent\textbf{Downstream actions.}
To test generalizability to recognizing semantically similar actions, we recommend evaluating the two within-set granularities of Gym-99 \ie, FX-S1 and UB-S1. 
Both of these subsets have high semantic similarity between actions, with many methods currently struggling to generalize to both of these subsets. Both are almost inversely correlated to Kinetics-400 performance, as can be seen in  \cref{proposed-benchmark}.
There is also a significant gap between CNNs and most transformer methods for both  subsets, highlighting the potential for future works in this area. Moreover, there is also no correlation with domain shift subsets and sample subsets, making these complementary challenges to both.

\noindent\textbf{Downstream task.} 
To evaluate the task sensitivity, we recommend using repetition counting on UCF-101 and multi-label classification on Charades. Repetition counting on UCF-101 highlights different strengths of action recognition as it allows investigation of a model's ability to generalize to a task that requires more temporal understanding without measuring the impact of the domain. 
It is evident from the table~\cref{proposed-benchmark} that it is currently a very challenging task for transformer-based methods, especially video-text models. 

We also recommend multi-label classification on Charades as it is currently challenging for video-only self-supervised models (both CNNs and transformers) and allows the combination of domain and task shift to be investigated.  

Overall, we hope that all the proposed subsets can act as challenging setups to holistically evaluate video representation learning methods for different downstream factors in addition to standard Kinetics-400 action recognition. We hope the future works can use the updated  \textbf{SEVERE-benchmark++} to evaluate the generalization capability of all types of video representation learning methods. 
\section{Observations}
We hope that our study and resulting benchmark provide a helpful insight for future research to design novel self-supervised and video-text methods for generalizable video representation learning. From the main results and benchmark results in~\cref{proposed-benchmark}, we make the following observations: 

\paragraph{No Clear Winner} 
Although different methods stand out in different downstream factors for all three pre-training types
there is no clear winner that achieves the best performance on all downstream settings.  

\paragraph{CNN vs. Transformer} While current research has converged on transformers for self-supervised (and other types of) video understanding, transformers are not the best across the different downstream factors. Transformers are more robust to domain and task shift than CNNs, however they underperform CNN-based approaches on fine-grained actions and when a limited number of finetuning samples are available. 

\paragraph{Video-only vs Video-text} From ~\cref{proposed-benchmark} we observe that despite using significantly smaller pre-training datasets, 240K in Kinetics-400 vs. 2M-25M pairs, video-only transformers are better or on par with video-text models. The best video-text model across downstream factors is UMT, which trains with a video-only masked encoder alongside the video-text data. While video-text models learn to connect vision and language modalities, our results suggest that this connection does not result in a better video representation. 
One of the reasons for this could be that video-text datasets are spatially focused~\cite{doughty2024locomotion}.  

\paragraph{Supervised vs Self-supervised} For the CNNs, supervised pre-training is dominant across all sensitivity factors, mainly when the number of downstream samples available is limited and when there is a change in both the downstream domain and the downstream task. For transformers, supervised pre-training significantly lags behind the best self-supervised methods and supervised CNN pre-training. 

\paragraph{Invariance Hurts Generalizability} Learning certain temporal invariances may prevent generalizability to temporal or fine-grained benchmarks. This is evident from GDT’s performance
on SS-v2 and UB-S1. These benchmarks require distinction between actions
such as moving something left vs. moving something right in SS-v3 and giant circle forwards vs. giant circle backwards in UB-S1. The invariance to
temporal reversal learned by GDT impacts its ability to recognize such actions.  For CNN-based methods, contrastive methods that explicitly encourage features to be distinct across the temporal dimension  \ie, GDT, TCLR, and RSPNet, tend to perform well.

\paragraph{Motion in pre-training}
Learning from synthetic motions improves the generalization performance as evidenced by CNN-based methods CtP and Tubelet-Contrast. Tubelet-Contrast is the best-performing method on several aspects of SEVERE, including downstream actions. In video-text pre-training we also see that synthetic motions can be useful with  LocoMotion's performance being on par with VindLU in ~\cref{proposed-benchmark} despite using 10 times less data.  Using synthetic motions in the pre-training objective requires the network to learn from moving pixels/objects rather than static background information of the videos, thus resulting in more generalizable representations.  This suggests that learning from such synthetic data or real datasets with more motions may be a way forward for self-supervised video representation learning, potentially in combination with other approaches such as masked autoencoding.  


\paragraph{Masked Autoencoding}
We observe that masked autoencoders that reconstruct high-level semantics instead of raw pixels obtain the best generalization. This includes HOG features used by MME and DINO features by SIGMA.


\section{Limitations }

While our study provides a comprehensive evaluation of the generalization capabilities of modern video representation learning methods, it is important to recognize several limitations. 

\noindent\textbf{pre-training Scope.}
We fix the pre-training dataset to Kinetics-400 for CNNs and video-only transformers and rely on publicly available pre-trained models. While this enables consistent comparison across methods and architectures, it introduces variability in the exact pre-training configurations, such as data augmentations, optimization strategies, or training duration, that may affect downstream performance.  Also, the size of the Kinetics-400 dataset is smaller than other available pre-training datasets~\cite{carreira2018short,carreira2019short,wang2023internvid}, which leaves the comparison on a much larger scale as an open question for future work. 

Moreover, due to a lack of a standard dataset in video-text pre-training, we evaluate models with different pre-training datasets. An ideal comparison would be to pre-train all methods with the same video-text pairs. One can further use the video-only data from such pairs to compare with CNN and video-only transformers for a perfect comparison. However, such an approach is computationally very expensive and not feasible with our available resources, thus we leave it for future works. 

\noindent\textbf{Model Architecture.}
For CNN-based methods, we use a fixed R(2+1)D-18 backbone to ensure a fair comparison with the category. While this choice reflects the standard in many prior works, it may constrain the full potential of some methods—especially on more complex datasets like EPIC-Kitchens-100 or AVA. For transformer-based methods, we evaluate with a ViT-B backbone, however, its common to train video-only and video-text models with larger backbone like ViT-L and ViT-H. A more detailed architectural study remains an open direction.

\noindent\textbf{Task Diversity.}
We have considered a selection of various video understanding tasks centered around human actions. However, many more video understanding tasks could be explored, such as human-centric tasks such as action anticipation~\cite{EPIC-100-arxiv}, as well as non-human-centric tasks such as animal behavior analysis \cite{animal1,animal2,animal3}, multi-object tracking~\cite{pedersen20203d}, visual grounding~\cite{animal3} and surgical video analysis~\cite{surgical}.

\noindent\textbf{Modality and Pretext Task Bias.}
While we include both video-only and video-text transformer models, our analysis is grounded in the downstream performance and does not fully disentangle the contribution of different pretext tasks, modalities, or training objectives. As many video SSL and video-text methods increasingly rely on cross-modal pre-training (e.g., with language, audio), a deeper investigation into which modalities drive generalization and how they interact under distribution shift is necessary.

\section*{Data Availability}
The pre-trained models and datasets used in this work are publicly available. We will release the models and code after acceptance. 

\section*{Acknowledgments}
This work is supported by the KAUST Center of Excellence for Generative AI under award number 5940. The computational resources are provided by IBEX, which is managed by the Supercomputing Core Laboratory at KAUST.

\begin{appendices}

\section{}
\label{sec:expt-details}

\subsection{Downstream Domain}
\label{app:domain-shift-expt}
In \cref{sec:factor_1} we investigate to what extent self-supervised methods learn features applicable to action recognition in any domain. Here we explain the datasets, splits and training details we use to do this.

\medskip
\noindent\textbf{Datasets} We report our experiments on the following datasets:\\
\textit{UCF-101} \cite{UCF-101-arxiv} is currently one of the most widely used datasets for evaluating video self-supervised learning models. It consists of YouTube videos from a set of 101 coarse-grained classes with a high overlap with actions in Kinetics-400. We use the first standard split proposed in the original paper \cite{UCF-101-arxiv} containing 9,537 training and  3,783 testing samples for the 101 action classes.\\
\textit{NTU-60}: \cite{NTU-60-arxiv} consists of daily human actions captured in a controlled lab setting with a fixed number actors. Although it has some overlap with Kinetics-400 actions, it is quite different visually due to the setting. We use the cross-subject protocol proposed in \cite{NTU-60-arxiv} to split the data into 40,320 training and 16,560 testing samples for 60 action classes.\\
\textit{Gym-99}. We use FineGym version $v1.0$ \cite{Gym-99-arxiv} which is a dataset of fine-grained actions constructed from  recorded gymnastic competitions. We use the Gym 99 subset which contains 99 action classes with 20,484 and 8,521 samples in the train and test sets respectively.\\
\textit{SS-v2}: \cite{SS-v2-arxiv} is a crowdsourced collection of first-person videos aimed to
instill
common-sense understanding. It differs significantly with respect to Kinetics-400 in terms of visual appearance and point-of-view. We use the original dataset splits from \cite{SS-v2-arxiv} containing 168,913  training and 24,777 testing samples for 174 action classes.\\
\textit{EPIC-Kitchens-100}: \cite{EPIC-100-arxiv} is a large-scale egocentric dataset consisting of daily actions performed in a kitchen. It has annotations for verbs (97) and nouns (300) and the action is defined a tuple of these. Like SS-v2, EK-100 also differs significantly from Kinetics-400 in terms of visual appearance and point-of-view. We use standard splits from \cite{EPIC-100-arxiv} containing 67,217 samples in training set and 9,668 in the validation set. We only aim to recognize the 97 verb classes. 
\begin{table*}[t]
\captionsetup{font=footnotesize,skip=1mm}
\centering
\caption{\textbf{Finetuning and Linear Evaluation details for CNN-based methods} on various downstream datasets. Learning rate is scheduled using a multip-step scheduler with $\gamma = 0.1$ at corresponding steps for each dataset. We train all the models with the same hyperparameters for the corresponding dataset.}
\resizebox{\textwidth}{!}{%
\begin{tabular}{lccccccccc}
\toprule
\multicolumn{1}{l}{\multirow{3}{*}{\textbf{Dataset}}} &
  \multicolumn{4}{c}{\textbf{Finetuning}} & 
  \multicolumn{1}{c}{} &
  \multicolumn{4}{c}{\textbf{Linear Evaluation}} \\
 \cmidrule{2-5} \cmidrule{7-10}
 & Batch Size & Learning rate & Epochs & Steps & & Batch Size & Learning rate & Epochs &  Steps \\
\midrule
UCF-101 &  32&  0.0001  & 160 & [60,100,140] & & 64&  0.01 & 100 & [40,80] \\
NTU-60 &   32&  0.0001 & 180 &  [90, 140, 160]  & &  64&  0.01 & 120 & [40,80,100] \\
Gym-99 &   32&  0.0001 & 160 & [60,100,140] & & 64 & 0.01 & 120 & [40,80,100] \\
SS-v2 &    32&  0.0001 & 45  & [25, 35, 40] & & 64&  0.01 & 40 & [20,30] \\
EK-100 &   32&  0.0025 & 30  & [20, 25] & & 32 &  0.0025 & 30 & [20, 25] \\
K-400   & 32 & 0.0001 & 50 & [25, 34, 45] & & 64&  0.01 & 40 & [10,20,30] \\
\bottomrule
\end{tabular}%
}
\label{tab:downstream_domain_training}
\end{table*}

\noindent\textbf{Training Details}
In the initial hyperparameter search, we perform a grid search over various finetuning settings with learning rates between 0.1 - 0.00001, varying total training epochs, data augmentations, and schedulers. We choose the optimal hyperparameters based on the performance of the pre-training models on the validation sets of each dataset for each downstream task. We do this for all three types of categories.

For \textbf{CNNs}, during finetuning and linear evaluation, we sample a random clip from each video of 32 frames with standard augmentations \ie, a random multi-scale crop of size 112 x 112, random horizontal flipping, and color jittering. We train with the Adam optimizer. The learning rates, scheduling and total number of epochs vary across datasets and are shown in \cref{tab:downstream_domain_training}. Each model is trained with the same hyperparameters for the corresponding dataset.  For inference, we use 10 linearly spaced clips of 32 frames each. For each frame, we take a center crop which is resized to 112x112 pixels. To calculate the action class prediction of a video, we take the mean of the predictions from each clip and report top-1 accuracy.

\begin{table*}[t]
    \centering
    \small
    \caption{\textbf{Finetuning details for Video-only transformer-based methods } on various downstream datasets. We train all the models with the same hyperparameters for the corresponding dataset.}
     \begin{tabular}{l|cccccc}
         \hline
         config & K400 & UCF-101 & NTU-60 & GYM-99 & SS-V2 & EK-100 \\
         \toprule
         optimizer & \multicolumn{6}{c}{AdamW} \\
         base learning rate & \multicolumn{4}{c}{1.0e-3} & {5.0e-4} & {5.0e-4} \\
         weight decay & \multicolumn{6}{c}{0.05} \\
         optimizer momentum & \multicolumn{6}{c}{$\beta_1,\beta_2=0.9,0.999$} \\
         layer-wise lr decay\cite{layer_wise} & \multicolumn{6}{c}{0.75} \\
         batch size & \multicolumn{6}{c}{16} \\
         learning rate schedule & \multicolumn{6}{c}{cosine decay} \\
         warmup epochs & \multicolumn{6}{c}{5} \\
         training epochs & 75 & 100 & 100 & 100 & 40 & 100 \\
         flip augmentation & \emph{yes} & \emph{yes} & \emph{yes} & \emph{yes} & \emph{no} & \emph{yes} \\
         RandAug~\cite{cubuk2019randaugment} & \multicolumn{6}{c}{(9,0.5)} \\
         label smoothing\cite{szegedy2015rethinking} & \multicolumn{6}{c}{0.1} \\
         mixup~\cite{zhang2018mixup} & \multicolumn{6}{c}{0.8} \\
         cutmix~\cite{yun2019cutmix} & \multicolumn{6}{c}{1.0} \\
         drop path & \multicolumn{6}{c}{0.1} \\
         \bottomrule
    \end{tabular}
    \label{tab:downstream_domain_training_video_trans}
    
\end{table*}

\begin{table*}[h]
    \centering
    \small
    \setlength{\tabcolsep}{4pt}
    \caption{\textbf{Linear Evaluation details for Video-only transformer-based methods } on various downstream datasets. We train all the models with the same hyperparameters for the corresponding dataset.}
     \begin{tabular}{l|cccccc}
     \hline
    config & K-400 & UCF-101 & NTU-60 & SS-v2 & GYM-99 & EK-100 \\
    
         \toprule
         optimizer & \multicolumn{6}{c}{AdamW} \\
         base learning rate & \multicolumn{6}{c}{1.e-3} \\
         weight decay & \multicolumn{6}{c}{0.05} \\
         optimizer momentum & \multicolumn{6}{c}{$\beta_1,\beta_2=0.9,0.999$} \\
         layer-wise lr decay~\cite{layer_wise} & \multicolumn{6}{c}{0.75} \\
         batch size & \multicolumn{6}{c}{128}\\
         learning rate schedule & \multicolumn{6}{c}{cosine decay} \\
         training epochs & 30 & 100 & 100& 50 &100 &100 \\
         flip augmentation & \emph{yes} & \emph{yes} & \emph{yes} & \emph{no} & \emph{yes} & \emph{yes} \\
         \bottomrule
    \end{tabular}
    \label{tab:downstream_domain_training_video_trans_linear}
\end{table*}

For \textbf{video-only transformers}, during the finetuning and linear evaluation stage, we adopt dense sampling as in \cite{tong2022videomae} to sample a random multi-scale crop of size 224 x 224 x 16 with standard augmentations like random horizontal flip, RandAug, mixup, and cutmix. For evaluation, all models follow a consistent inference strategy, using 5 x 3 or 3 x 2 crops.
The learning rates, scheduling, and total number of epochs vary across datasets and are shown in \cref{tab:downstream_domain_training_video_trans} and \cref{tab:downstream_domain_training_video_trans_linear}. Each model is trained with the same hyperparameters for the corresponding dataset as before.

For \textbf{video-text transformers}, during the finetuning and linear evaluation stage, we again adopt dense sampling as in \cite{tong2022videomae} to sample a random multi-scale crop of size 224 x 224 x 8 with standard augmentations like random horizontal flip, RandAug, mixup, and cutmix. We use positional embedding interpolation to use  8 frames during the downstream tasks. For evaluation, all models follow a consistent inference strategy, using 5 x 3 or 3 x 2 crops.
The learning rates, scheduling, and total number of epochs vary across datasets and are shown in \cref{tab:downstream_domain_training_video_text_trans} and \cref{tab:downstream_domain_training_video_text_trans_linear}. Each model is trained with the same hyperparameters for the corresponding dataset as before.

\begin{table*}[t]
    \centering
    \small
    \caption{\textbf{Finetuning details for Video-text transformer-based methods } on various downstream datasets.}
     \begin{tabular}{l|cccccc}
         \hline
         config & K400 & UCF-101 & NTU-60 & GYM-99 & SS-V2 & EK-100 \\
         \toprule
         optimizer & \multicolumn{6}{c}{AdamW} \\
         base learning rate & \multicolumn{6}{c}{5.0e-4, 5.0e-5 } \\
         weight decay & \multicolumn{6}{c}{0.05} \\
         optimizer momentum & \multicolumn{6}{c}{$\beta_1,\beta_2=0.9,0.999$} \\
         layer-wise lr decay\cite{layer_wise} & \multicolumn{6}{c}{0.75} \\
         batch size & \multicolumn{6}{c}{32,128} \\
         learning rate schedule & \multicolumn{6}{c}{cosine decay} \\
         warmup epochs & \multicolumn{6}{c}{5} \\
         training epochs & 50 & 100 & 100 & 100 & 30,40 & 100 \\
         flip augmentation & \emph{yes} & \emph{yes} & \emph{yes} & \emph{yes} & \emph{no} & \emph{yes} \\
         RandAug~\cite{cubuk2019randaugment} & \multicolumn{6}{c}{(9,0.5)} \\
         label smoothing\cite{szegedy2015rethinking} & \multicolumn{6}{c}{0.1} \\
         mixup~\cite{zhang2018mixup} & \multicolumn{6}{c}{0.8} \\
         cutmix~\cite{yun2019cutmix} & \multicolumn{6}{c}{1.0} \\
         drop path & \multicolumn{6}{c}{0.1} \\
         \bottomrule
    \end{tabular}
    \label{tab:downstream_domain_training_video_text_trans}
    
\end{table*}

\begin{table*}[t]
    \centering
    \small
    \setlength{\tabcolsep}{4pt}
    \caption{\textbf{Linear Evaluation details for Video-text transformer-based methods } on various downstream datasets. }
     \begin{tabular}{l|cccccc}
     \hline
    config & K-400 & UCF-101 & NTU-60 &  GYM-99 & SS-v2  & EK-100 \\
         \toprule
         optimizer & \multicolumn{6}{c}{AdamW} \\
         base learning rate & \multicolumn{6}{c}{1.e-2,1.e-3} \\
         weight decay & \multicolumn{6}{c}{0.05} \\
         optimizer momentum & \multicolumn{6}{c}{$\beta_1,\beta_2=0.9,0.999$} \\
         layer-wise lr decay~\cite{layer_wise} & \multicolumn{6}{c}{0.75} \\
         batch size & \multicolumn{6}{c}{128}\\
         learning rate schedule & \multicolumn{6}{c}{cosine decay} \\
         training epochs & 50 & 100 & 100& 100 &50 &100 \\
         flip augmentation & \emph{yes} & \emph{yes} & \emph{yes} & \emph{yes} & \emph{no} & \emph{yes} \\
         \bottomrule
    \end{tabular}
    \label{tab:downstream_domain_training_video_text_trans_linear}
\end{table*}

\subsection{Downstream Samples}
In \cref{sec:factor_2} we measure how sensitive current video self-supervised models are to the amount of downstream samples. We do this by varying the size of the training data starting from 1000 examples and doubling it until we reach the full train set. We use the same data splits as in the downstream domain experiments, explained in \cref{app:domain-shift-expt}, and sample a subset of video clips from the respective train sets. We use the same random subset across the different models to make the comparison fair. For each dataset, we follow the same training and testing procedure as the full datasets setting in downstream domain experiments~\cref{app:domain-shift-expt} and \cref{tab:downstream_domain_training}, \cref{tab:downstream_domain_training_video_trans} and \cref{tab:downstream_domain_training_video_text_trans}. 

\subsection{Downstream Actions}
In \cref{sec:factor_3} we measure how benchmark-sensitive current video self-supervised models are to downstream actions. We do so by measuring performance on different subsets, defined in the FineGym dataset~\cite{Gym-99-arxiv}, which have increasing semantic similarity. We provide the details of Gym-99, Gym-288 and the four different subsets we use of Gym-99 below: 

\noindent\textbf{Gym-99} consists of 29k video clips of 99 different actions across the four different gymnastic events in FineGym: Vault, Floor Exercise, Balance Beam and Uneven Bars. This is a relatively balanced subset of the full FineGym dataset with all actions having more than 80 occurrences. There are a total 20.5k training videos and 8.5k testing videos.

\noindent\textbf{Vault} is a subset of Gym 99 containing 1.5k videos of the 6 actions from the Vault event. 
The training split contains 1.0k examples and the testing split contains 0.5k examples.

\noindent\textbf{Floor} contains actions in the Floor Exercise event from Gym-99. It consists of 7.5k instances of over 35 actions with a split of 5.3k for training and 2.2k for testing. 

\noindent\textbf{FX-S1} is a subset of actions of leaps, jumps and hops from the Floor event in Gym-99. This subset of 11 actions 
contains a total of 2.6k video clips with 1.9k for training and 0.7k for testing.

\noindent\textbf{UB-S1} contains 5k videos of 15 actions from the Uneven Bars event with a split of 3.5k for training and 1.5k for testing. The actions consist of different types of circles around the bars. 

\noindent\textbf{Gym-288} is a long-tailed version of Gym 99 containing 32k videos with 22.6K training and 9.6K testing samples. It adds 189 infrequent classes to the 99 classes in Gym 99, where actions can have as little as 1 or 2 instances in training. This results in a total of 288 action classes from the four different gymnastic events. 

We follow the same training and evaluation procedure as that for the full dataset  finetuning of Gym-99 in downstream domain training, as shown in \cref{tab:downstream_domain_training}, \cref{tab:downstream_domain_training_video_trans} and \cref{tab:downstream_domain_training_video_text_trans}.

\subsection{Downstream Tasks}
In \cref{sec:factor_4} we investigate how sensitive self-supervised methods are to the downstream task and whether they generalize beyond action recognition. We provide details of the experimental setup used for each task below.

\noindent\textbf{Spatio-temporal Action Detection.}  
The goal of this task is to localize actors in video clips with bounding boxes across both spatial and temporal dimensions, and to classify their actions. We conduct experiments on the UCF101-24 benchmark, a subset of UCF-101, which provides bounding box annotations for 3,207 videos across 24 action classes.
For CNN models, we follow the implementation of K{\"{o}}p{\"{u}}kl{\"{u}} \etal~\cite{yowo}, using only a 3D-CNN branch for spatio-temporal action detection. The 3D backbone is initialized with a self-supervised pre-trained R(2+1)D-18 model.
For video-only and video-text transformer models, we adopt ViT-B as the backbone, pre-trained in a self-supervised manner.
During training:
\begin{itemize}
  \item For \textbf{CNN} and \textbf{video-only} models, we sample 16-frame clips and apply standard augmentations, including horizontal flipping, random scaling, and random spatial cropping.
  \item For \textbf{video-text} models, the number of input frames varies by architecture.
\end{itemize}
The training setups are as follows:
\begin{itemize}
  \item \textbf{CNN models} are trained with the Adam optimizer, an initial learning rate of $1\times10^{-4}$, weight decay of $5\times10^{-4}$, and a batch size of 64, for a total of 12 epochs. The learning rate follows a multi-step schedule with $\gamma{=}0.5$ at epochs [4, 6, 8, 10].
  \item \textbf{Video-only models} use the AdamW optimizer with an initial learning rate of $5\times10^{-4}$, weight decay of 0.05, and a batch size of 16, for 50 epochs, using a cosine decay learning rate schedule.
  \item \textbf{Video-text models} adopt architecture-specific settings; \eg VindLU uses a learning rate of $1\times10^{-4}$, while VICLIP uses $4\times10^{-6}$.
\end{itemize}
During evaluation, we follow \cite{yowo} and report video-level mean Average Precision (video-mAP) across all action classes.

 \noindent\textbf{Repetition counting}. The goal 
 of the this task is to estimate the number of times an action repeats in a video clip. We use the UCFRep benchmark proposed by Zhang \etal \cite{rep_counting}, which is a subset of UCF-101. The dataset consists of 526 videos with 3,506 repetition number annotations. From the annotated videos, 2M sequences of 32 or 16  frames and spatial size 112x112 or 224x224 are constructed which are  used as the input.  We use the implementation from the original benchmark \cite{rep_counting} with pre-trained R(2+1)D-18 and ViT-B models as the backbone networks. Each model is trained for 100 epochs with a batch size of 32 using the Adam optimizer with a fixed learning rate of 0.00005. For testing, we follow the protocol from \cite{rep_counting} and report mean counting error.
    
\noindent \textbf{Arrow-of-time}. The goal of this task is to predict the direction (forward of backward) of the video. We closely follow the setup used by Ghodrati \etal \cite{arrow_of_time}. The full UCF-101 dataset is used with two versions of each video, one normal and one reversed. During training, for each video, we sample 8 frames linearly with a random offset, with a batch size of 12 and 112x112 or 224x224 center crops, number of epochs 10, learning rate of $1e^{-5}$. We do not use any augmentations. During testing, we sample 8 frames linearly. We report top-1 binary classification accuracy.

\noindent \textbf{Temporal Action Localization on UCF domain}  Temporal Action Localization (TAL)~\cite{liu2025opentad,zhao2023re2tal,zhang2022actionformer} is a task focused on identifying action categories within a video and precisely determining the start and end times of each action instance. For task shift within domain, we use THUMOS-14~\cite{jiang2014thumos} which is from the same domain as UCF-101.
The THUMOS14~\cite{jiang2014thumos} dataset consists of 413 untrimmed videos spanning 20 action categories. It is split into a validation set with 200 videos and a test set with 213 videos. In line with standard practice~\cite{zhang2022actionformer,zhao2023re2tal,liu2025opentad}, we train on the validation set and evaluate performance on the test set.
In our study, we evaluated all methods on THUMOS-14~\cite{jiang2014thumos}. For this, we used pre-trained models from each method to extract spatio-temporal features and finetuned them using ActionFormer~\cite{zhang2022actionformer}, implemented within the OpenTAD framework~\cite{liu2025opentad}.
Following standard practice in the TAL community, we report average mean Average Precision (mAP) over multiple temporal Intersection over Union (tIoU) thresholds: 5 tIoU values ([0.3, 0.4, 0.5, 0.6, 0.7]).

\noindent\textbf{Multi-label classification on Charades}. Charades \cite{charades-sigurdsson:hal-01418216} is made up of videos of people recording everyday activities at their homes.  
Videos in Charades are longer than the other datasets we use and the goal is to recognize multiple different actions in each video. A per-class sigmoid output is used for multi-class prediction.
We use the implementation of Feichtenhofer \etal \cite{large-scale-feichtenhofer2021large}\footnote{\href{https://github.com/facebookresearch/SlowFast}{https://github.com/facebookresearch/SlowFast}} with the R(2+1)D-18 and ViT-B backbone. 
During training, we use 32 or 16  frames with a sampling rate of 8. Since this task requires longer temporal context, we observe that using more frames with higher sampling rate is beneficial. We use a spatial crop of 112x112 or 224x224 and augmentations such as random short-side scaling, random spatial crop and horizontal flip. We train for 57 epochs in total with a batch size of 16 and a learning rate of 0.0375 with multi-step scheduler with $\gamma = 0.1$ at epochs [41, 49]. During testing, following \cite{large-scale-feichtenhofer2021large}, we spatio-temporally max-pool predictions over 10 clips for a single video. We report mean average precision (mAP) across classes.

\noindent\textbf{Action Detection on AVA.}  
The AVA dataset~\cite{AVA-Gu_2018_CVPR} consists of video clips extracted from movies, with bounding box annotations for temporally fine-grained action classes. We use version v2.2 for spatio-temporal action detection. The goal is to detect and classify actions from proposals generated by an off-the-shelf person detector.
For CNN models, we follow the implementation of~\cite{large-scale-feichtenhofer2021large} and adopt R(2+1)D-18 as the backbone. During training, we sample 32-frame clips at a rate of 2, apply a spatial crop of $112 \times 112$, and use data augmentations including random short-side scaling, random spatial cropping, and horizontal flipping. The model is trained for 20 epochs using a multi-step learning rate scheduler (initial learning rate = 0.1, decay factor $\gamma = 0.1$ at epochs [10, 15]) and a batch size of 32. During testing, we use a single 8-frame clip, sampled at a rate of 8 and centered in the video.

For video-only models, we follow \cite{tong2022videomae} with ViT-B as the backbone. The training input consists of 64-frame clips sampled at a rate of 4, with a multi-scale crop of $224 \times 224$ and horizontal flipping as data augmentation. Models are trained for 50 epochs using a cosine decay learning rate scheduler (initial learning rate = 0.0005) and a batch size of 16. For testing, we use a single 16-frame clip sampled at a rate of 4, centered in the video.

For video-text models we adopt \cite{tong2022videomae} with ViT-B as the backbone.. We adopt the AdamW optimizer with $(\beta_1, \beta_2) = (0.9, 0.999)$ and a weight decay of 0.05. A cosine learning rate scheduler is applied across all video-text models, while the specific learning rates, the number of training epochs and the number of input frames vary depending on the model. We report mean Average Precision (mAP) across all action classes.

\noindent\textbf{Temporal Action Localization on ActivityNet.} 
Temporal Action Localization (TAL)~\cite{liu2025opentad,zhao2023re2tal,zhang2022actionformer} is a task focused on identifying action categories within a video and precisely determining the start and end times of each action instance. It requires the model to comprehend both spatial semantics within individual frames and temporal dynamics across sequences of frames to capture the progression of actions. In our study, we evaluated all methods on ActivityNet-v1.3~\cite{caba2015activitynet}. 
ActivityNet-v1.3~\cite{caba2015activitynet} is a large-scale benchmark covering 200 activity classes across approximately 20,000 videos, totaling over 600 hours of content. The dataset is split into training, validation, and test sets, with 10,024, 4,926, and 5,044 videos respectively. Consistent with prior works~\cite{zhang2022actionformer,zhao2023re2tal,liu2025opentad}, we train our models using the training split and report results on the validation set. For this, we used pre-trained models from each method to extract spatio-temporal features and finetuned them using ActionFormer~\cite{zhang2022actionformer}, implemented within the OpenTAD framework~\cite{liu2025opentad}.
Following standard practice in the TAL community, we report average mean Average Precision (mAP) over multiple temporal Intersection over Union (tIoU) thresholds: 10 tIoU values ([0.5, 0.55, 0.6, 0.65, 0.7, 0.75, 0.8, 0.85, 0.9, 0.95]).

\subsection{Downstream Dataset Attributes}
\label{sec:video-datasets}

\begin{figure}
    \centering
    \includegraphics[width=\linewidth]{media/radar_datasets.pdf}
    \includegraphics[width=\linewidth]{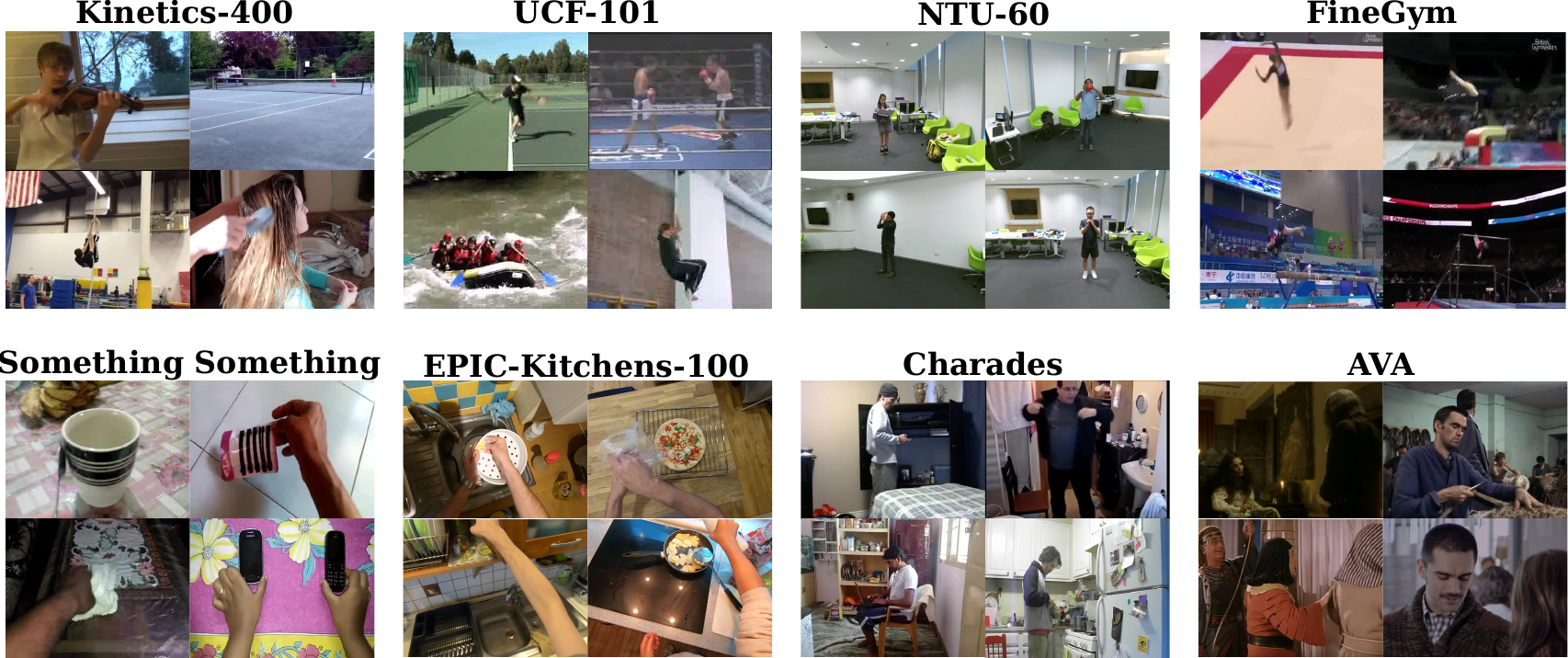}
    \caption{Example video frames from the Kinetics-400 pre-training dataset and some downstream datasets we consider. Note the differences in the capture settings and point-of-view across these datasets.}
    \label{domain_frames_appendix}
\end{figure}

\begin{figure}
    \centering
    \begin{tabular}{@{}c@{}c}
        \includegraphics[width=0.8\linewidth]{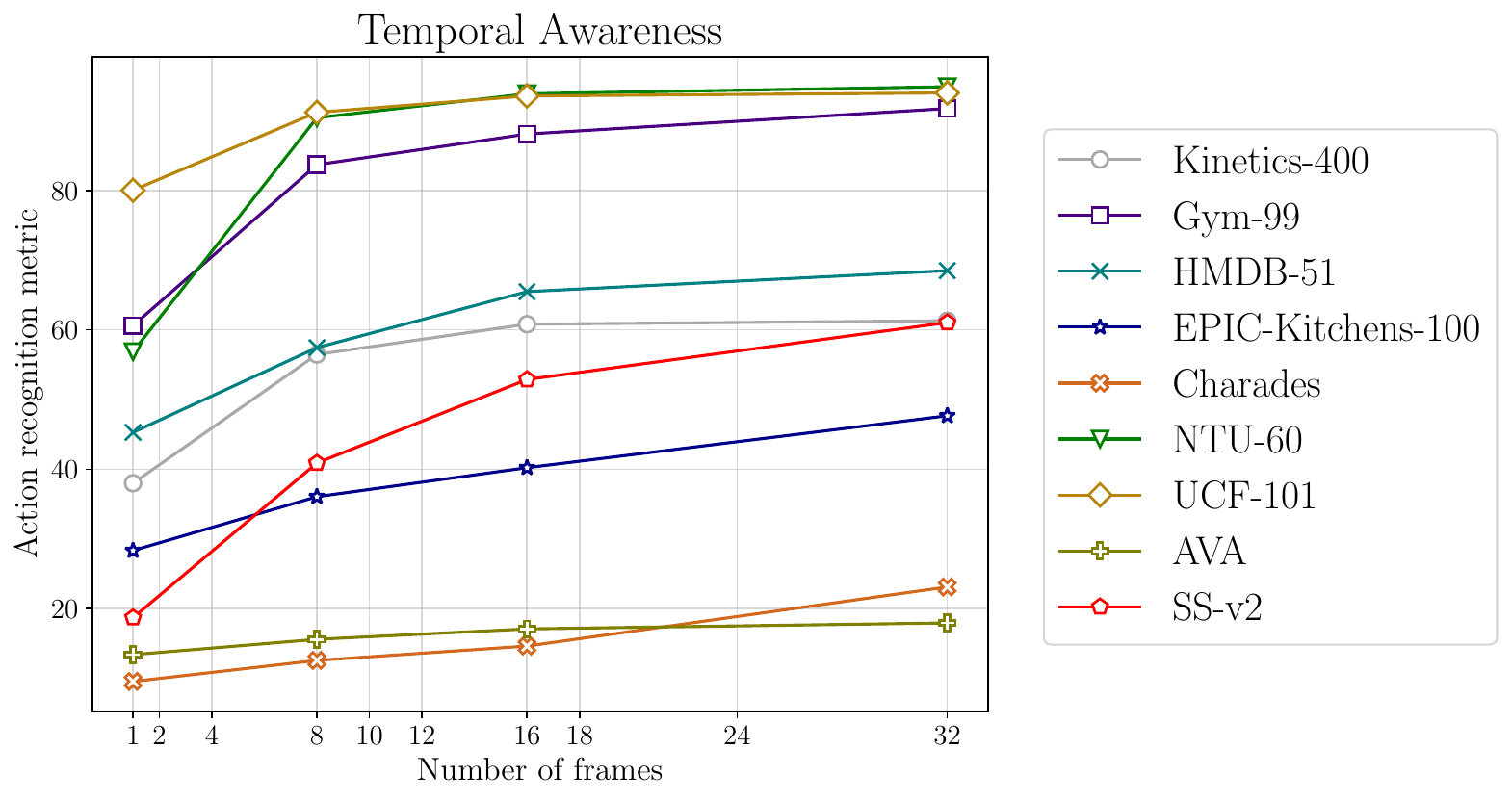} 
    \end{tabular}
    \caption{\small \textbf{Temporal awareness}. Illustrating the effect of temporal awareness (increasing temporal-context)  on the action recognition performance using a standard 3D-CNN for different action datasets.
    }
    \label{fig:action_temporality-intradataset}
\end{figure}

\begin{table*}[]
\captionsetup{font=footnotesize,skip=1mm}
\centering
\caption{\textbf{Pre-training differences of our evaluated CNN-based self-supervised methods.} All models are pre-trained with the same R(2+1)D-18 backbone and Kinetics-400 dataset, there are differences in how many epochs they were trained for, the batch size and number of frames they use and the spatial and temporal augmentations they are encouraged to be invariant to.}
\resizebox{\textwidth}{!}{%
\begin{tabular}{lllllccccccccccc}
\toprule
\multicolumn{1}{l}{\multirow{3}{*}{\textbf{Method}}} &
  &  & & && 
  \multicolumn{6}{c}{\textbf{Spatial Augmentations}} &
  \multicolumn{1}{c}{} &
  \multicolumn{3}{c}{\textbf{Temporal Augmentations}} \\
 \cmidrule{7-12} \cmidrule{14-16}
\multicolumn{1}{c}{} &  Extra & Epochs & Batch & Num & 
  \multicolumn{1}{c}{} &
  \multicolumn{1}{c}{Random} &
  \multicolumn{1}{c}{Horiz.} &
  \multicolumn{1}{c}{Grayscale} &
  \multicolumn{1}{c}{Color} &
  \multicolumn{1}{c}{Gaussian} &
  \multicolumn{1}{c}{Scaling} &
  \multicolumn{1}{c}{} &
  \multicolumn{1}{c}{Shift} &
  \multicolumn{1}{c}{Reversal} &
  \multicolumn{1}{c}{Speed} \\
  & Modality & & Size & Frames  & & Crop & Flip & & Jitter & Blur & \\
\midrule
MoCo & & 200 & 128 & 16 &  & \ding{51} & \ding{51} & \ding{51} & \ding{51} & & & &\ding{51}& & \\
SeLaVi &
  Audio & 200 & 1024 & 30 &  &
  \ding{51} &
  \ding{51} &
   &
   &
   &
   &
   &
   &
   &
    \\
  VideoMoCo & &
  200 & 128 & 32 &  & 
  \ding{51} &
  \ding{51} &
  \ding{51} &
  \ding{51} &
  &
  &
   &
  &
  & \\
  Pretext-Contrast &
   & 200 & 16 & 16 &  &
  \ding{51} &
  \ding{51} &
  \ding{51} &
  \ding{51} &
  \ding{51} &
   &
   &
  \ding{51} &
   &
    \\
RSPNet &
   & 200 & 64 & 16 & &
  \ding{51} &
   &
   &
  \ding{51} &
  \ding{51} &
   &
   &
  \ding{51} &
   &
  \ding{51} \\
  AVID-CMA &
  Audio & 400 & 256 & 16 &  &
  \ding{51} &
  \ding{51} &
   &
  \ding{51} &
   &
  \ding{51} &
   &
   &
 \\
 CtP &
   & 90 & 32 & 16  
 \\
TCLR &
   & 100 & 40 & 16&  &
  \ding{51} &
  \ding{51} &
  \ding{51} &
  \ding{51} &
   &
  \ding{51} &
   &
   &
   & \\
GDT &
  Audio & 100 & 512 & 30 &  &
  \ding{51} &
  \ding{51} &
   &\ding{51}
   &
   &
   &
   &
   &
  \ding{51} &
   \\
   \midrule
   Supervised & & 45 & 32 & 16 &  &
   \ding{51} & \ding{51} 
   &
   & 
   &
   &
   &
   &
   \ding{51}\\
\bottomrule
\end{tabular}%
}
\label{tab:cnn_pretrain}
\end{table*}

\begin{table*}[h]
    \centering
    \small
    \caption{\textbf{Pre-training differences of our evaluated Video-only transformer-based methods.} All models are pre-trained with the same ViT-B backbone and Kinetics-400 dataset; there are differences in how many epochs they were trained for, the batch size, the masking type, and the reconstruction targets.}
    \begin{tabular}{l|ccccccc}
         config & VideoMAE & EVEREST & MVD & MGMAE & MME & MGM & SIGMA \\
         \toprule
         optimizer & \multicolumn{7}{c}{AdamW} \\
         base learning rate & {1.5e-4} & {3e-4} & {1.5e-4} & {1e-3} & {1.5e-4} & {1.5e-4} & {1.5e-4} \\
         weight decay & \multicolumn{7}{c}{0.05} \\
         optimizer momentum & \multicolumn{7}{c}{$\beta_1,\beta_2=0.9,0.95$} \\
         batch size  & {256} & {1024} & {1024} & {384} & {768} & {512} & {640} \\
         learning rate schedule & \multicolumn{7}{c}{cosine decay} \\
         warmup epochs & {40} & {40} & {40} & {20} & {40} & {40} & {40} \\
         flip augmentation &  \multicolumn{7}{c}{\emph{yes}} \\
         augmentation & \multicolumn{7}{c}{MultiScaleCrop} \\
         Epochs & {800} & {200} & {1600+400} & {200} & {800} & {800} & {800}\\
         Masking Guidance & random & random & random & motion & random &  motion & random\\
         Reconstruction Target & Pixels & Pixels & Pixels & Pixels & HOG Features & Pixels & DINO Features\\
    \toprule
    \end{tabular}
    \label{tab:vo_trans_pretrain}
\end{table*}

\begin{table*}[h]
    \centering
    \small
    \caption{\textbf{Pre-training differences of our evaluated Video-text transformer-based methods } }
    \begin{tabular}{l|cccc}
         config & Locomotion & VindLU & UMT & VICLIP \\
         \toprule
         optimizer & \multicolumn{4}{c}{AdamW} \\
         base learning rate & {1e-4} & {1e-4} & {1e-4} & {2e-4} \\
         weight decay & {0.02} & {0.02} & {0.02} & {0.2} \\
         optimizer momentum & \multicolumn{3}{c}{$\beta_1,\beta_2=0.9,0.999$} & {$\beta_1,\beta_2=0.9,0.98$} \\
         batch size & {160} & {2048} & {4096} & {512} \\
         learning rate schedule & \multicolumn{4}{c}{cosine decay} \\
         warmup epochs & {1} & {1} & {1} & {0.5} \\
         flip augmentation & \multicolumn{4}{c}{\emph{yes}} \\
         augmentation & {random resize, crop} & {random resize, crop} & {MultiScaleCrop} & {random resize, crop} \\
         Epochs & {5} & {10} & {10} & {10} \\
         \# frames & 4 & 4 & 8 & 8 \\
         dataset & WebVid-2.5M~\cite{bain2021frozen} & WebVid-25M~\cite{bain2021frozen} & K700\cite{carreira2019short}+WebVid-25M\cite{bain2021frozen} & InternVid~\cite{wang2023internvid} \\
    \toprule
    \end{tabular}
    \label{tab:vt_trans_pretrain}
\end{table*}

We define several attributes in \cref{subsec:domain-shift} in order to characterize differences in domain between the downstream datasets and the Kinetics-400 pre-training dataset in \cref{fig:radar}. 
The attributes \textit{Point-of-view} and \textit{Environment} are defined qualitatively based on the contents of the target dataset. Radar plot and sample examples of videos from each of the datasets are shown in \cref{domain_frames_appendix}. We can see that FineGym \cite{Gym-99-arxiv} consists of videos of Olympic gymnastic events. Thus, we label it as \textit{stadium} for environment and \textit{third-person} for point-of-view. On the radar plots, we order the environment in descending order of variability contained in a given dataset. Kinetics-400 is placed near the origin as it has much higher variability than NTU-60, for example, which is captured in a controlled lab setting. \textit{Action length} is the average duration of the actions in each of the datasets. 

We quantify \textit{temporal awareness} as the minimum number of frames (temporal context) required to best recognize the action. We do this by finetuning R(2+1)D with weights initialized from supervised pre-training on Kinetics-400 and we denote temporal awareness ($\tau$) as:
\begin{equation}
    \tau = \arg\min_{t \in\{1, 2, ..., N\}}\left[ \left(100 \times \frac{f_{t+1} - f_{t}}{f_{t}}\right) < \alpha \right]
\end{equation}
where $\alpha$ is chosen to be $1$. This means $\tau$ indicates the number of frames after which relative improvement in performance is lesser than $\alpha$, \ie when the performance has plateaued. \cref{fig:action_temporality-intradataset} shows the top-1 action recognition performance against increasing number of frames for each of our downstream datasets. We use bilinear interpolation to estimate performance at given number of frames beyond those that we experiments with. For example, using our method to compute temporal awareness, the performance for UCF-101 plateaus at 7 frames while that for EK-100 plateaus at 32 frames indicating that EK-100 needs much larger temporal context for recognition while UCF-101 may suffice with a shorter temporal context.

\textit{Label overlap} is the amount of actions which are present in both the downstream dataset and the pre-training dataset (Kinetics-400). We quantify this by matching identical actions as well as manually checking for reworded versions of the same action class. 
For example, ``head massage'' in UCF-101 has a corresponding action ``massaging person's head'' in Kinetics-400. In NTU-60 action class ``brushing teeth'' has a matching action ``brushing teeth'' in Kinetics-400. 

\subsection{Pre-training Differences}
\label{app:pretraining_differences}

We list the pre-training differences of all three types of methods in \cref{tab:cnn_pretrain}, \cref{tab:vo_trans_pretrain}, and \cref{tab:vt_trans_pretrain}

\end{appendices}

\bibliography{sn-bibliography}
\end{document}